\documentclass[preprint,11pt]{elsarticle}

\makeatletter
\def\ps@pprintTitle{%
	\let\@oddhead\@empty
	\let\@evenhead\@empty
	\def\@oddfoot{\centerline{\thepage}}%
	\let\@evenfoot\@oddfoot}
\makeatother

\usepackage{cite}
\usepackage{amsmath,amssymb,amsfonts}
\usepackage{hyperref}
\usepackage[linesnumbered,ruled,vlined]{algorithm2e}
\usepackage{listings}
\usepackage{graphicx}
\usepackage{textcomp}
\usepackage{xcolor}
\usepackage{bm}
\usepackage{tabularx}
\usepackage{multirow}
\usepackage{makecell}
\usepackage{verbatim}

\newcommand{\C}{{\mathbb C}}
\newcommand{\R}{{\mathbb R}}

\newcommand{\be}{\begin{equation}}
	\newcommand{\ee}{\end{equation}}
\newcommand{\ba}{\begin{array}}
	\newcommand{\ea}{\end{array}}
\newcommand{\baa}{\left[\begin{array}}
	\newcommand{\eaa}{\end{array}\right]}

\newcommand{\beqa}{\begin{eqnarray}}
	\newcommand{\eeqa}{\end{eqnarray}}
\newcommand{\bt}{\begin{tabular}}
	\newcommand{\et}{\end{tabular}}

\newcommand{\bi}{\begin{itemize}}
	\newcommand{\ei}{\end{itemize}}
\newcommand{\bc}{\begin{center}}
	\newcommand{\ec}{\end{center}}

\newtheorem{Defi}{Definition}

\newtheorem{prop}[Defi]{Proposition}
\newtheorem{remark}[Defi]{Remark}
\newtheorem{example}[Defi]{Example}

\def\QED{\hfill \mbox{\rule[0pt]{1.5ex}{1.5ex}}}

\begin{document}
	
	\begin{frontmatter}
		\title{Haar-Laplacian for directed graphs\tnoteref{ack_Asydil}
		}
		\tnotetext[ack_Asydil]{This work was supported by a grant of the Ministry of Research,
			Innovation and Digitization, CNCS - UEFISCDI,
			project number PN-III-P4-PCE-2021-0154, within PNCDI III.}
		
		\author{Theodor-Adrian Badea, Bogdan Dumitrescu
		}
		
		\address{Department of Automatic Control and Computers \\
			National University of Science and Technology Politehnica Bucharest, Romania \\
			Emails: theodor.badea@upb.ro, bogdan.dumitrescu@upb.ro.}

\begin{abstract}
This paper introduces a novel Laplacian matrix aiming to enable the construction of spectral convolutional networks and to extend the signal processing applications for directed graphs. Our proposal is inspired by a Haar-like transformation and produces a Hermitian matrix which is not only in one-to-one relation with the adjacency matrix, preserving both direction and weight information, but also enjoys desirable additional properties like scaling robustness, sensitivity, continuity, and directionality. We take a theoretical standpoint and support the conformity of our approach with spectral graph theory. Then, we address two use cases: graph learning (by introducing HaarNet, a spectral graph convolutional network built with our Haar-Laplacian) and graph signal processing.
We show that our approach gives better results in applications like weight prediction and denoising on directed graphs.
\end{abstract}


\end{frontmatter}


\section{Introduction}
L{et} $\mathcal{G} = \left(\mathcal{V}, \; \mathcal{E}\right)$, with $|\mathcal{V}| = N$, be a graph; $\mathcal{V}$ and $\mathcal{E}$ are the sets of vertices and edges, respectively.
We assume that the graph is simple: there is at most an edge $(u,v)$, for any vertices $u,v \in \mathcal{V}$.
If the graph is undirected, then the edges $(u,v)$ and $(v,u)$ are identical.
An edge $(u, v) \in \mathcal{E}$ has a weight $w_{uv} \in \R$, $w_{uv} \ne 0$;
the weights can have negative values.
The adjacency matrix $A$ of $\mathcal{G}$ is defined by
\be
	a_{uv} = \begin{cases}
		w_{uv}, & \text{if $(u, v) \in \mathcal{E}$} \\
		0, & \text{otherwise}.
	\end{cases}
\label{adjm}
\ee
A multigraph can be accommodated to this framework by defining $w_{uv}$
as the sum of the weights of all edges between $u$ and $v$.

We assume that the graph does not have self-loops, hence $a_{uu}=0$, $\forall u \in \mathcal{V}$.
We also assume that the graph is connected; there are no isolated vertices.

Let $D$ be the diagonal matrix
\be
D = \mathrm{diag} (|A| \cdot \bm{1}),
\label{diag_degree}
\ee
where $|A|$ is the matrix whose elements are $|a_{uv}|$
and $\bm{1}$ is a vector with all elements equal to 1.
When all weights are equal to one, this is the degree matrix:
$d_{uu}$ is the number of edges containing vertex $u$.

The Laplacian associated with the graph is
\be
L = D - A
\label{lapl}
\ee
and the symmetrically normalized Laplacian is 
\be
\bar{L} = D^{-1/2} L D^{-1/2}
= I_N - D^{-1/2} A D^{-1/2}.
\label{norm_lapl}
\ee
If the graph is undirected with nonnegative weights, the Laplacian enjoys nice spectral properties, which proved useful in graph signal processing \citep{SaMo13GSP,MSM20GSP,Isufi24graph}.
In particular,  both Laplacians \eqref{lapl} and \eqref{norm_lapl}
are symmetric and positive semidefinite;
the eigenvalues of the normalized Laplacian \eqref{norm_lapl} lie in the interval $[0,2]$.
The case of negative weights in undirected graphs is covered in \citep{Kunegis10signed}.

{\em Contribution.}
Here we propose a new construction, called Haar-Laplacian, based on the symmetrized and skew-symmetrized adjacency matrices associated with a directed graph.
Like other Laplacians for directed graphs \citep{Shubin94magnet,Zh21_magnet,Fior23sigmanet}, the Haar-Laplacian is a Hermitian matrix and thus has real eigenvalues and orthogonal eigenvectors.
We show that the Haar-Laplacian enjoys some properties that former Laplacians do not share.
Moreover, we use the Haar-Laplacian in two types of applications. For graph learning, we introduce a spectral graph convolutional network called HaarNet and show its behavior for several instances of the link prediction problem in directed graphs; the best results are obtained for weight prediction, where the one-to-one relation of the Haar-Laplacian with the adjacency matrix gives it an edge over the existing approaches that do not have this property.
We also illustrate the benefits of a variant of the Haar-Laplacian,
with differences only in the diagonal, but with a sound frequency domain interpretation, in graph signal processing tasks such as denoising
A related work is \citep{BaDu24dualgcn}, where we have used the symmetrized and skew-symmetrized adjacency matrices for building separate blocks in an autoencoder network dedicated to anomaly detection; no connection with the Laplacian was made; no results, theoretical or experimental, are replicated in this paper.

{\em Contents.}
In Sec. \ref{sec:context}, we review the state-of-the-art in the topic of Laplacians for directed graphs, from two perspectives, those of graph learning and graph signal processing.
In Sec. \ref{sec:haarlap}, we introduce the Haar-Laplacian and discuss its properties, showing its advantages over existing Laplacians.
We also present its properties for simple graphs.
Section \ref{sec:learning} is dedicated to the description of the graph learning problems that we tackle and to the presentation of the solution based on the Haar-Laplacian, the HaarNet network.
Section \ref{sec:results} contains the numerical results obtained for learning (three types of link prediction problems and a node classification problem) and signal processing (denoising).
Section \ref{sec:concl} concludes the paper.

\section{Context}
\label{sec:context}

The core of our discussion is a key concept in spectral graph theory, namely the Graph Fourier Transform (GFT).
The transformation itself leverages the eigenvectors of the Laplacian to perform signal processing on graphs. 
Essentially, the GFT enables the analysis and manipulation of signals defined over graph nodes, transforming them from the vertex domain to a frequency domain that reflects the underlying graph structure. 
This notion is employed in both the learning and signal processing matters that we will discuss in further detail throughout this paper.

Considering a graph signal $x \in \R^N$ on the graph $\mathcal{G}$, the Fourier transform and its inverse are defined as
\begin{equation}
	\label{fourier}
	\hat{x} = U^T x, \; \; x = U\hat{x}.
\end{equation}
where $U = \left[u_0, ..., u_{N-1}\right]$ is the matrix of eigenvectors (Fourier basis) corresponding to the diagonal matrix $\Lambda = \mathrm{diag}\left(\left[\lambda_0, ..., \lambda_{N-1}\right]\right)$ of eigenvalues (frequencies) of the Laplacian.
As mentioned in the previous section, if $\mathcal{G}$ is undirected with nonnegative weights, then it enjoys nice spectral properties.
It has a full set of real and nonnegative eigenvalues with corresponding orthonormal eigenvectors, thus highlighting the meaning of $\Lambda$ and $U$ in the GFT definition.

However, when the graph is directed, the spectrum of the Laplacian is complex and
it is hard to associate it with meaningful spectral graph operations.
Several Laplacians have been proposed, starting with that of the symmetrized
adjacency matrix
\be
A_s = \frac{1}{2} \left( A + A^T \right).
\label{sym_adj}
\ee
Using it in \eqref{lapl} or \eqref{norm_lapl} gives symmetric positive semidefinite Laplacians.
However, such a Laplacian only captures the average weight between two nodes;
the differences are lost.
It is in fact a brutal replacement of a directed graph with an undirected one,
all information about orientation being lost.
Such a Laplacian was used in \citep{KiWe16}, in the context of a spectral
graph convolutional network (GCN),
with good results in classification.
When only connectivity matters, this Laplacian can be successfully used.

\subsection{Learning perspective}
\label{sec:context_learning}

We start by discussing the viewpoint taken in learning applications
related to directed graphs.

Perhaps the most prominent association of the Laplacian within the frames of learning applications is with spectral graph convolutional networks (GCNs), which extend the convolutional neural networks (CNNs) to graph-structured data. 
Graphs lack an appropriate translation operation; therefore, the convolution in the vertex domain is unfeasible. 
However, the convolution operation on graphs is enabled in the spectral domain by means of the graph Fourier transform (GFT) \citep{Bruna2014spectral}.
Then, in a manner similar to the classical signal processing, the convolution of two graph signals $x$ and $y$ in the spectral domain turns into a multiplication:
\begin{equation}
	\label{eq:convolution}
	x \ast_\mathcal{G} y = U \left( \left( U^T x \right) \odot \left(U^T y \right) \right),
\end{equation}
where $U$ is the GFT matrix in \eqref{fourier} and $\odot$ is the element-wise (Hadamard) product. 

This approach allows GCNs to extend the concept of spatial locality to the unique and irregular structure of graphs. 
In plain terms, a layer of a GCN is a convolution between a graph signal and a learnable filter $g_\theta$. 
A rudimentary $g_\theta$ could be a nonparametric filter, but such a filter would not be localized \citep{defferrard2017convolutional}.
Moreover, the convolution itself is a computationally intensive matrix multiplication, and dealing with large graphs makes the Laplacian eigendecomposition a prohibitive operation.
Hence, to address these issues, it was suggested to make use of a recursively-formulated polynomial evaluated in the Laplacian as a parametrization for the filter.
The recurrence of Chebyshev polynomials can be exploited \citep{hammond2011wavelets} to approximate $g_\theta$ as a truncated expansion, which further allows writing the filtering procedure as
\begin{equation}
	\label{eq:filtering}
	g_\theta \ast_\mathcal{G} x = g_\theta\left(\bar{L}\right)x = \sum_{k = 0}^{K-1} \theta_k T_k \left(\frac{2}{\lambda_{max}} \bar{L} - I_N\right) x,
\end{equation}
where $\theta_k$ are learnable filter coefficients and $\lambda_{max}$ is the largest eigenvalue of the normalized Laplacian $\bar{L}$ \eqref{norm_lapl}. In \citep{KiWe16}, the simplification to a model that performs first-order local smoothing is proposed, which is computationally efficient and sufficient for many tasks. Additionally, $\lambda_{max} = 2$ is employed as an approximation, this way eliminating the need to actually compute the largest eigenvalue.

To address the asymmetry issues of directed graphs, several approaches to extend the spectral methods have been proposed.
The most interesting build a Laplacian by using Hermitian matrices, which have real eigenvalues.
The magnetic Laplacian \citep{Shubin94magnet,Zh21_magnet} is such a solution.
(An early proposal \citep{Chung05lapl} has several computational disadvantages and its normalized version can be dense.) 
Denoting
\be
D_s = \mathrm{diag}(|A_s| \cdot \bm{1}),
\label{Ds}
\ee
the magnetic Laplacian is
\be
L_m = D_s - H,
\label{magn_lapl}
\ee
with
\be
H = A_s \odot \mathrm{exp} (i \bm{\Theta}),
\label{magn_adj}
\ee
where $\odot$ is the element-wise (Hadamard) product,
\be
\theta_{uv} = 2 \pi q (a_{uv} - a_{vu}),
\label{phase_magnet}
\ee
and $q \ge 0$ is a parameter.
So, $\bm{\Theta}$ acts as a phase matrix, being zero when the graph is undirected;
the direction of the edges is visible in $\bm{\Theta}$;
if $(u,v) \in \mathcal{E}$, but $(v,u) \not \in \mathcal{E}$, then $\mathrm{Im}(h_{uv})>0$;
in the opposite case, where $(v,u) \in \mathcal{E}$, but $(u,v) \not \in \mathcal{E}$,
then $\mathrm{Im}(h_{uv})<0$.
Taking $q=0.25$, if $|a_{uv}| \le 1$, it follows that $\theta_{uv} \in [-\pi,\pi]$,
which ensures the unicity of the phase.
The magnetic Laplacian \eqref{magn_lapl} is positive semidefinite.
The same property applies to its normalized version
\be
\bar{L}_m = I_N - D_s^{-1/2} H D_s^{-1/2}.
\label{norm_magn_lapl}
\ee

A drawback of the magnetic Laplacian is its lack of robustness to scaling.
When multiplying $A$ with a constant $c$, the real part of \eqref{magn_lapl}
scales correctly; however, the phase is multiplied by $c$,
so it may go outside the interval $[-\pi,\pi]$ and thus change its meaning.
For example, take $a_{uv}=1$, $a_{vu}=0$ and so
$h_{uv} = 0.5(\cos(\pi/2) + i \sin(\pi/2)) = 0.5i$.
Then, for the matrix $4A$ we would obtain $h_{uv} = 2(\cos(2\pi) + i \sin(2\pi))=2$;
this value is the same as that of an adjacency matrix with $a_{uv}=a_{vu}=2$,
which has two antiparallel edges and could thus be associated with symmetry.
Similarly, for the matrix $3A$, it results that $h_{uv}=1.5(\cos(3\pi/2) + i \sin(3\pi/2))=-1.5i$;
hence, one may say that the edge is reversed (or the sign of the weight is flipped).

So, working with the magnetic Laplacian implies special care in choosing the value
of the parameter $q$.
For example, in \citep{He22msgnn}, where the magnetic Laplacian is generalized for negative weights, the value
$q = 1/2 \max |a_{uv} - a_{vu}|$ is taken, ensuring the non-ambiguity of the phase.
In \citep{Zhang21mgc}, the value of $q$ is chosen by appealing to Johnson's algorithm for
detecting directed cycles.
However, these methods are efficient when data properties are known;
the behavior for new data may not be robust.

A more general cure was proposed in \citep{Fior23sigmanet}, where the
sign-magnetic Laplacian was introduced.
It replaces \eqref{magn_adj} with
\be
H = A_s \odot \left( \bm{1} \cdot \bm{1}^T - \mathrm{sgn}(|A-A^T|) 
+ i \cdot \mathrm{sgn}(|A|-|A^T|)\right),
\label{sigmag_adj}
\ee
where sgn is the sign function.
Indeed, in this case scaling works as it should: magnitude is modified, but phase is not.
In all the above examples with $a_{uv}=1$, $a_{vu}=0$, it results that $h_{uv}$ is purely imaginary,
with positive imaginary part.
Inserting \eqref{sigmag_adj} in \eqref{magn_lapl} and \eqref{norm_magn_lapl}
gives positive definite matrices.

However, while the sign-magnetic Laplacian captures perfectly the orientation
of the edges, it is not devoid of criticism.
Essentially, we would like that different graphs have different Laplacians,
a fact that is true for undirected graphs;
however, this is not true for the Laplacians in this section.
In particular, the sign-magnetic Laplacian fails when both edges $(u,v)$ and $(v,u)$ exist.
It is easy to see in \eqref{sigmag_adj} that the same value $h_{uv}$ is obtained
by taking, for example, $a_{uv}=10000$ and $a_{vu}=1$ in one case and
$a_{uv}=5001$ and $a_{vu}=5000$ in the other;
in both cases, it results that $h_{uv}=5000.5 i$; 
so, there is no sensitivity to variations that may be huge.
Also, the sign-magnetic Laplacian is not continuous;
if $a_{uv}=1$ and $a_{vu}=1$, then $h_{uv}=1$; if $a_{uv}=1+\epsilon$ and $a_{vu}=1-\epsilon$, with $\epsilon > 0$
then $h_{uv}=i$;
there is a palpable discontinuity even though the change of weights may be minor.
Finally, the sign-magnetic Laplacian is zero when $A$ is skew-symmetric, since $A_s=0$.

We thus appreciate that the sign-magnetic Laplacian is not necessarily the
best option for graphs with pairs of antiparallel edges, like those modeling
money transfers (it matters how imbalanced are the transfers between two accounts,
not only the average transferred amount)
or trust networks; if trust is evaluated on a scale from $-10$ to $10$,
having $a_{uv}=10$ and $a_{vu}=-10$ means that $u$ totally trusts $v$, but $v$ has no
confidence at all in $u$; this can be a stab-in-the-back situation;
while $a_{uv}=1$ and $a_{vu}=-1$ might mean that $u$ and $v$ barely know each other;
however, both cases are coded with the same value, $0$, in the sign-magnetic Laplacian (as well as in the magnetic Laplacian).

Other Laplacians in the same vein \citep{He22msgnn,Singh22_signed_GNN,Ko23spectral} share some
of the above weaknesses.
In \citep{Ko23spectral}, the phase is coded such that all edges orientations and
weight signs have distinct values in the Laplacian;
however, like in the sign-magnetic Laplacian, there is no sensitivity to weight
disparity in antiparallel edges.
Although not used in the spectral context, the Hermitian adjacency matrix \citep{Liu2015HAM,GuMo17HAM},
which may be seen as a simplified version of \eqref{magn_adj},
has been recently used in learning \citep{Ke24HAM} and could be used for building a Laplacian.

In \citep{Fior24quat}, the antiparallel edges (digons) are finally addressed and
a solution is proposed by the use of quaternions, as a generalization of both the classical and the sign-magnetic Laplacians.
The Laplacian is no longer a Hermitian complex matrix, but a Hermitian quaternion.
Different graphs have different Laplacians; however, the mapping is not surjective:
there are Hermitian quaternions that do not correspond to graphs.
So, there is finally a distinction between all situations, but it is obtained
with an excess of means.
Also, the mapping from the adjacency matrix to the quaternion Laplacian is not continuous;
the situation is similar to that of the sign-magnetic Laplacian.

Another track of research, that we do not follow here, is based on Laplacians
that are symmetric although the graph is directed.
An example, based on a personalized PageRank approach, is \citep{Tong2020inception}.
Also aside our interests are approaches that only approximate the Laplacian \citep{Tong2021contrastive}.

\subsection{Signal processing perspective}

We present now the viewpoint taken in graph signal processing (GSP) applications,
where the main theoretical notions associated to a Laplacian
are those of Graph Fourier Transform (GFT) and graph frequencies \citep{SaMo14GSP}.
Since the topic is very generous and a wealth of information can be found in overview papers \citep{SaMo13GSP,MSM20GSP,Isufi24graph} (the latter two emphasizing the case of directed graphs),
we only mention the main approaches.

Random walks on directed graphs lead to a Laplacian construction \citep{LiZh12};
the same work also defines the notion of degree of asymmetry, which measures the skewness of the
Laplacian (and adjacency matrix).
A more detailed analysis is presented in \citep{Sevi23}.
The GFT may be obtained by diagonalization, which is not always possible.
Like the GFT via the Jordan decomposition \citep{Singh16_GFT_Jordan}, computation can be
affected by numerical instability.
Also, the GFT is not orthogonal, which is always desirable for a transform.

So, again, Hermitian Laplacians come into play, with their convenient spectral properties.
For example, the magnetic Laplacian was proposed for GSP in \citep{Furutani19_magnet}.

In \citep{CCS23}, the singular value decomposition (SVD) of \eqref{lapl} is used as basis
for a GFT. The GFT is real, but its size is double, since two orthogonal matrices are involved.
The singular values of the Laplacian are the frequencies.
The left and right singular vectors get the meaning of frequency components.
A somewhat related method, in the sense that SVD has a central role, uses
the polar decomposition \citep{Kwak24polar}.

Optimization based methods try to build an orthogonal transformation that
minimizes the directed total variation \citep{Sardellitti17_GFT}
or the spectral dispersion \citep{Shafipour18_GFT} of the graph.
A more general view of the GFT is given in \citep{GON18}.

A comparison of these Laplacians would be quite ample, since theoretically they all have
some useful properties and the range of applications is large.
So, we will proceed by introducing our proposed Laplacian.

\section{The Haar-Laplacian}
\label{sec:haarlap}

\subsection{Definition and properties}

The shortcomings discussed in Section \ref{sec:context_learning} are eliminated by our proposal.
Let
\be
A_a = \frac{1}{2} \left( A - A^T \right)
\label{asym_adj}
\ee
be the skew-symmetric matrix naturally generated from the adjacency matrix $A$.

\begin{remark} \rm
	Building the symmetrized \eqref{sym_adj} and skew-symmetrized \eqref{asym_adj}
	matrices amounts to a Haar-like transformation (spatial instead of temporal) of the weights.
	Together, these matrices contain the same information as $A$
	(the "inverse transform" is $A=A_s+A_a$);
	the symmetric part contains the averages ("low frequency" information)
	and the skew-symmetric part contains the differences ("high frequency").
	So, the situation is similar with that of a Haar filter bank, although no
	explicit downsampling is involved; the number of independent elements is the same
	in $A$ as in $A_s$ and $A_a$ together, and the transformation is linear.
	\QED
\end{remark}

\begin{Defi} \rm
	\label{def:haarlap}
	Consider the Hermitian matrix
	\be
	H_h = A_s + i A_a
	\label{haarlap_adj}
	\ee
	and the diagonal matrix
	\be
	D_h = \mathrm{diag}(|H_h| \cdot \bm{1}).
	\label{diag_h}
	\ee
	We define the {\em Haar-Laplacian}
	\be
	L_h = D_h - H_h
	\label{haar_lapl}
	\ee
	and the {\em normalized Haar-Laplacian}
	\be
	\bar{L}_h = I_N - D_h^{-1/2} H_h D_h^{-1/2}.
	\label{norm_haar_lapl}
	\ee
\end{Defi}

\begin{remark} \rm
	The elements of the Hermitian matrix \eqref{haarlap_adj} obey to
	\be
	|h_{uv}| = \sqrt{\frac{1}{2}(a_{uv}^2 + a_{vu}^2)}.
	\label{abs_huv}
	\ee
\end{remark}

\begin{prop} \rm
	\label{prop:posdef}
	The Haar-Laplacians \eqref{haar_lapl} and \eqref{norm_haar_lapl} are positive
	semidefinite and have real eigenvalues.
	The eigenvalues of the normalized Haar-Laplacian \eqref{norm_haar_lapl}
	lie in the interval $[0,2]$.
\end{prop}

{\em Proof.}
Both Haar-Laplacians are Hermitian matrices and hence have real eigenvalues.
The matrix
\[
\bar{H}_h = D_h^{-1/2} H_h D_h^{-1/2}
\]
is similar with
\[
D_h^{-1/2} (D_h^{-1/2} H_h D_h^{-1/2}) D_h^{1/2} = D_h^{-1} H_h.
\]
The sum of the absolute values of the (non-diagonal) elements on each row is
\[
\sum_{v=1, v\ne u} \frac{|h_{uv}|}{d_{uu}} \le 1,
\]
due to \eqref{diag_h}.
Since the diagonal elements of the matrix $\bar{H}_h$ are zero,
Gershgorin's circle theorem \citep{gerschgorin} says that its eigenvalues lie
inside the unit disk.
It results immediately that the eigenvalues of \eqref{norm_haar_lapl} are in $[0,2]$
and hence the matrix \eqref{norm_haar_lapl} is positive semidefinite.
Since
\[
L_h = D_h^{1/2} \bar{L}_h D_h^{1/2},
\]
the matrix \eqref{haar_lapl} is also positive semidefinite.
\QED

\begin{remark} \rm
	The properties discussed critically in Section \ref{sec:context_learning} for other Laplacians
	hold trivially for the Haar-Laplacian \eqref{haar_lapl}.
	Some of the properties follow from the fact that no information is lost
	from $A$ to $H_h$; the map $A \longrightarrow H_h$ is one-to-one;
	none of the previous Laplacians has this property. (Note that self-loops would allow several graphs to have the same Laplacians, which would be detrimental in learning problems.)
	
	{\em Scaling.}
	If $A$ is replaced by $cA$, then $H_h$ is replaced by $cH_h$.
	The ratio between the imaginary and real parts stays the same,
	so the phase is unchanged.
	
	{\em Sensitivity.}
	A change in $A$ will lead to a change in $H_h$.
	
	{\em Continuity.}
	The map $A \longrightarrow H_h$ is continuous.
	
	{\em Orientation.}
	The Haar-Laplacians clearly show the prevalent orientation of the edges.
	If $a_{uv} > a_{vu}$ (one of the weights may be zero, if the edge does not exist),
	then the signs of the real and imaginary parts of $h_{uv}$ show if the flow goes
	from $u$ to $v$ or viceversa and if the average flow is positive or negative.
	One can see also if only one edge exists: the real and imaginary parts
	are equal in absolute value. (This wasn't possible for the magnetic and sign-magnetic
	Laplacians.)
	\QED
\end{remark}

\begin{remark} \rm
	The idea of the proof of Prop. \ref{prop:posdef}, based on Gershgorin circles,
	can be used for most Laplacians mentioned in this paper; the proof is much shorter than existing ones.
	\QED
\end{remark}

\begin{remark} \rm
	The Haar-Laplacian is Hermitian, hence it has real eigenvalues.
	However, the eigenvectors are complex, in general.
	They form a unitary matrix.
	
	When the matrix $A$ is symmetric, $H_h=A_s$ is real and symmetric and
	its eigenvectors are real.
	The Haar-Laplacian \eqref{haar_lapl} is identical with the standard Laplacian \eqref{lapl} (and so is the magnetic Laplacian).
	
	When the matrix $A$ is skew-symmetric and so $A_s=0$,
	then $H_h = iA_a$ and its eigenvectors have real and imaginary parts
	with equal norm for all nonzero eigenvalues.
	This property is certainly known, but we did not find a proof.
	So, we give it below.
	Note that for skew-symmetric matrices, both magnetic and sign-magnetic Laplacians are zero.
	\QED
\end{remark}

\begin{prop} \rm
	Let $A_a$ be a real skew-symmetric matrix, $i\lambda$, with $\lambda \in \R$,
	one of its eigenvalues 
	(a skew-symmetric matrix has purely imaginary eigenvalues),
	and $x+iy$ the corresponding eigenvector.
	If $\lambda \ne 0$, then $\|x\|=\|y\|$.
	If $\lambda = 0$, then $y=0$.
\end{prop}

{\em Proof.}
Since $A_a (x+iy) = i \lambda (x+iy)$, it results that
$A_a x = - \lambda y$ and $A_a y = \lambda x$.
From the first equality, we can write
$|\lambda| \|y\|=\|A_a x\| \le \|A_a\| \|x\|$.
If $\lambda \ne 0$, taking into account that $|\lambda| \le \|A_a\|$,
it results that $\|y\| \ge \|x\|$.
Similarly, from the second inequality we obtain $\|x\|\ge\|y\|$.

If $\lambda=0$, then $Ax=Ay=0$, hence one can take $y=0$
(the eigenvector is naturally real).
\QED

\subsection{Frequency domain interpretation}
Let $z \in \C^N$ be a signal on the graph.
For an undirected graph with positive weights
(we assume $a_{uv}\ge 0$ for the discussion below),
a measure of the variability of a signal on the graph is
the total variation (TV)
\be
\text{TV}_2(z) = \frac{1}{2} \sum_{u,v=1}^N a_{uv} d(z_u,z_v)  = z^H L z,
\label{tv2}
\ee
where $d(z_u,z_v) = \left| z_u - z_v \right|^2$ and $L$ is the Laplacian \eqref{lapl}.
In GSP context, the eigenvalues of $L$ are named graph frequencies and the matrix
of eigenvectors is called GFT.
As discussed in \citep{SaMo14GSP} and other works, smaller eigenvalues are associated with
lower frequencies, since obviously TV is small in this case.
In particular, since $\bm{1}$ is the eigenvector corresponding to the eigenvalue $0$,
it results that $\text{TV}_2(\bm{1})=0$. 
Another way of expressing the squared difference between two signal values is
\be
d(z_u,z_v) = \baa{cc} z_u^* & z_v^* \eaa Q \baa{c} z_u \\ z_v \eaa.
\label{distQ}
\ee
Relation \eqref{tv2} is given by $Q = \baa{cc} 1 & -1 \\ -1 & 1 \eaa$.
Note the $Q$ is positive semidefinite; one of the eigenvalues is $0$ and the corresponding
eigenvector is $\bm{1}$. 

When the graph is directed, TV \eqref{tv2} takes into account only the symmetric part
of the adjacency matrix $A$.
To involve the antisymmetric part and thus make a connection with Hermitian Laplacians,
we can redefine \eqref{distQ} by using a complex matrix
\be
Q = \baa{cc} 1 & \alpha - i \beta \\ \alpha + i \beta & 1 \eaa,
\label{Qherm}
\ee
where $\alpha, \beta \in \R$ are parameters that can be chosen.
Since we want TV to be zero for a constant vector, the condition
$0 = \bm{1}^T Q \bm{1} = 2 + 2 \alpha$ gives $\alpha=-1$.
For such $\alpha$, the matrix $Q$ is no longer positive semidefinite if $\beta \ne 0$.
However, we can define a TV measure by using the absolute value of \eqref{tv2}:
\be
\text{TV}_Q(z) = \left| \frac{1}{2} \sum_{u,v=1}^N a_{uv} d(z_u,z_v) \right|,
\label{tvQ}
\ee
together with \eqref{distQ} and \eqref{Qherm}.
We can now make the connection with a slightly modified Haar-Laplacian.

\begin{prop} \rm
\label{p:haarD}
For a graph with positive weights, let us define the Laplacian
\be
\tilde{L}_h = D_s - H_h,
\label{haarlap_d}
\ee
where $D_s$ is defined in \eqref{Ds} and $H_h$ in \eqref{haarlap_adj}.
Note that this is the Haar-Laplacian with a different diagonal;
we will call \eqref{haarlap_d} the HaarD-Laplacian.
Then, for all $z \in \C^N$, the following equality holds:
\be
\text{TV}_Q(z) = \left| z^H \tilde{L}_h z \right|,
\label{haard_tv}
\ee
where $\text{TV}_Q$ is defined in \eqref{tvQ}, with $Q$ as in \eqref{Qherm} and
$\alpha=-1$, $\beta=1$.
\QED
\end{prop}

{\em Proof.}
Relation \eqref{haard_tv} can be proved directly, by expliciting the two sides.
Details are shown in Appendix A. 
\QED

\begin{remark} \rm
\label{rem:freq}
Let
\be
L_h = U \Lambda U^H, \ \tilde{L}_h = \tilde{U} \tilde{\Lambda} \tilde{U}^H
\label{eigLh}
\ee
be the eigenvalue decompositions of the Haar- \eqref{haar_lapl}
and HaarD-Laplacian \eqref{haarlap_d}, where
$\Lambda = \text{diag}( \lambda_1, \ldots, \lambda_N)$
and $U \in \C^{N \times N}$ is unitary and contains on columns the eigenvectors
$u_1, \ldots, u_N$;
$\tilde{\Lambda}$ and $\tilde{U}$ are defined similarly.

The relation \eqref{haard_tv} between total variation and the HaarD-Laplacian
suggests that, seen as graph frequencies, the eigenvalues of $\tilde{\Lambda}$ should
be ordered in increasing order of their absolute value and numbered such that
$|\lambda_1| \le \ldots \le |\lambda_N|$;
remind that they are not necessarily positive.
Lacking any other information, the eigenvalues of the Haar-Laplacian are numbered increasingly:
$\lambda_1 \le \ldots \le \lambda_N$;
they are all nonnegative.

Note that although $\bm{1}^T \tilde{L}_h \bm{1}=0$ (since $\text{TV}_Q(\bm{1})=0$ by
the choice of the matrix $Q$ in \eqref{Qherm}), the vector $\bm{1}$ is not necessarily
an eigenvector of the HaarD-Laplacian.
On the downside, in general, there are other vectors $z$ for which $\text{TV}_Q(z)=0$, although their elements are not equal.
So, the interpretation of low frequencies may be somewhat fuzzy.
However, at the other end of the spectrum, large frequencies are clearly related with
a high value of $\text{TV}_Q$ \eqref{tvQ}.
\QED
\end{remark}

\begin{remark} \rm
For a real signal $z \in \R^N$, relation \eqref{haard_tv} becomes
\be
\text{TV}_Q(z) = z^T (D_s - A_s) z \ge 0,
\ee
with $D_s$ from \eqref{Ds} and $A_s$ from \eqref{sym_adj}.
This is due to the fact that $z^T A_a z=0$ for a skew-symmetric matrix $A_a$.
The matrix $D_s-A_s$ is the Laplacian associated with the symmetrized adjacency matrix
and is positive semidefinite.
This is another way of saying that $\text{TV}_Q$ reduces to $\text{TV}_2$ for real signals.
\end{remark}

\begin{figure}[t]
	\centering
	\includegraphics[width=0.3\textwidth]{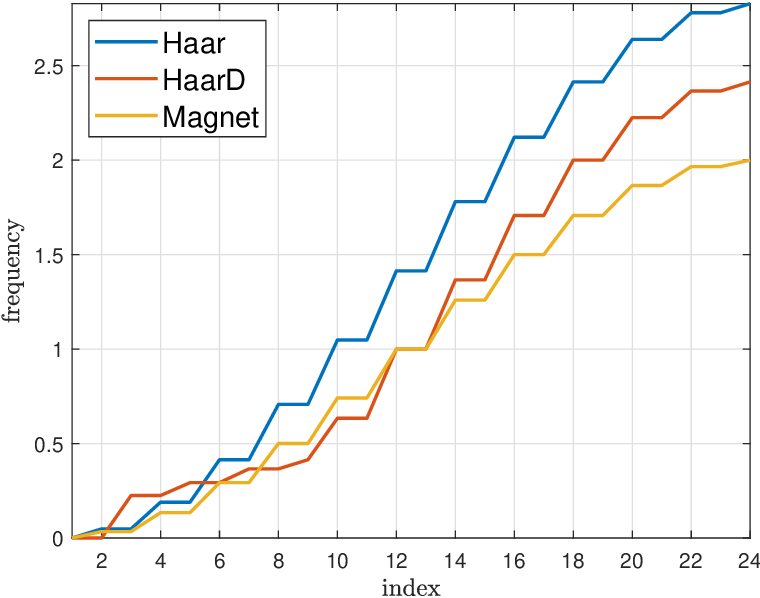} \\ 
	\includegraphics[width=0.3\textwidth]{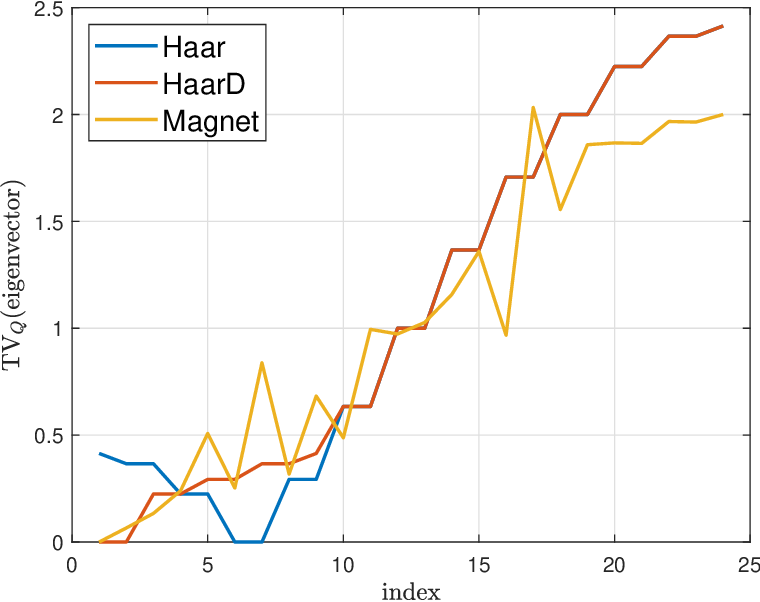}	 
	\caption{Top: sorted frequencies for Haar, HaarD, and magnetic ($q=1/4$) Laplacians
	for $N=24$. Bottom: total variation \eqref{tvQ} of the eigenvectors of the same Laplacians.}
	\label{fig:eig_circ}
\end{figure}

\begin{example} \rm	
	A testbed for the GFT is its form for the directed cycle graph, whose adjacency matrix is
	(here illustrated for $N=4$)
	\be
	S = \baa{cccc}
	0 & 1 & 0 & 0 \\
	0 & 0 & 1 & 0 \\
	0 & 0 & 0 & 1 \\
	1 & 0 & 0 & 0
	\eaa
	\label{shiftmat}
	\ee
	The corresponding Haar- and HaarD-Laplacians have the form
	\be
	\gamma I - \frac{1+i}{2}S - \frac{1-i}{2} S^T,
	\label{haarshift}
	\ee
	with $\gamma=\sqrt{2}$ for Haar and $\gamma=1$ for HaarD.
	Like \eqref{shiftmat}, the matrix \eqref{haarshift} is also circulant,
	hence the GFTs $U_h$ and $\tilde{U}_h$ in \eqref{eigLh}
	are permutations of the discrete Fourier transform matrix $W$.
	So, the Fourier transform of this graph is basically the same as
	in classical signal processing, like for the random walk Laplacian \citep{Sevi23}.
	However, the frequencies are different.
	Since $S$ in \eqref{shiftmat} is also diagonalized by $W$ and its eigenvalues are $\text{exp}(2\pi i j/n)$, $j=0:N-1$, the eigenvalues of the Laplacians \eqref{haarshift} are
	\be
	\gamma - \cos(2 \pi j/N) - \sin(2 \pi j/N), \ j=0:N-1.
	\label{eig_circ}
	\ee
	They take values in the interval $[\gamma - \sqrt{2}, \gamma+\sqrt{2}]$.
	Due to the symmetries of the function $\cos \xi + \sin \xi$ with respect to
	$\pi/4$ and $5\pi/4$, the $N$ values \eqref{eig_circ} are distinct when $N$ is
	not a multiple of 4; otherwise, excepting the smallest and the largest, they appear in pairs.
	
	

Figure \ref{fig:eig_circ} (top) shows the sorted frequencies (see Remark \ref{rem:freq})
of the Haar, HaarD and magnetic Laplacians (for $q=1/4$, one of the most popular choices)
for $N=24$.
The distribution of frequencies is similar.
Note that for $q=1/4$, the magnetic \citep{Zh21_magnet} and
sign-magnetic \citep{Fior23sigmanet} Laplacians are identical for the directed cycle graph.
For other values of $q \le 1/2$, they are generally different.

Figure \ref{fig:eig_circ} (bottom) shows the TV \eqref{tvQ} for the eigenvectors of the
above Laplacians. 
For HaarD, the curve is identical with that from the top figure, since
$\text{TV}_Q(\tilde{u}_j) = |\tilde{\lambda}_j|$, see \eqref{haard_tv},
and the eigenvalues are sorted in increasing order of absolute value.
For the Haar-Laplacian, the order is different; the low frequencies correspond
to low TV values, but not in increasing order; however, the high frequency eigenvectors
are the same as for the HaarD-Laplacian.
The TV for the magnetic Laplacian eigenvectors is generally increasing as eigenvalues
increase, but not monotonically; for other values of $N$ the behavior can be even
more chaotic.
\QED
\end{example}

\begin{figure*}
	\centering
	\includegraphics[width=0.3\textwidth]{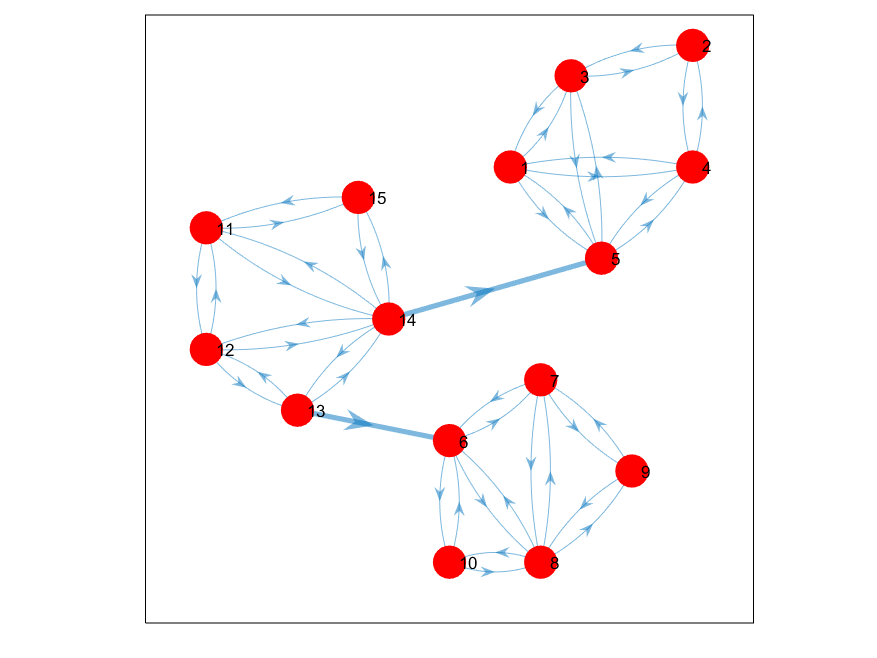}
	\includegraphics[width=0.3\textwidth]{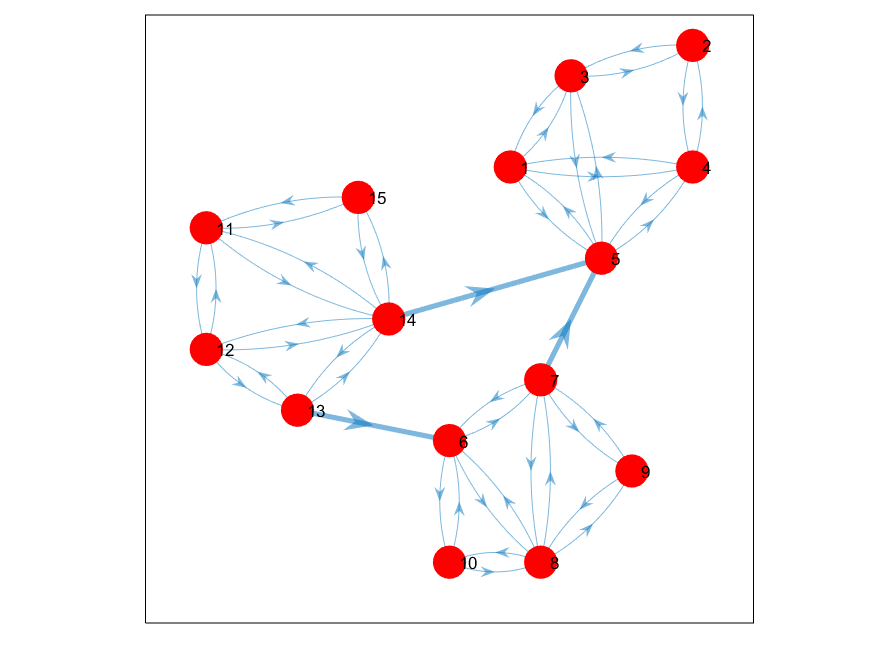}
	\includegraphics[width=0.3\textwidth]{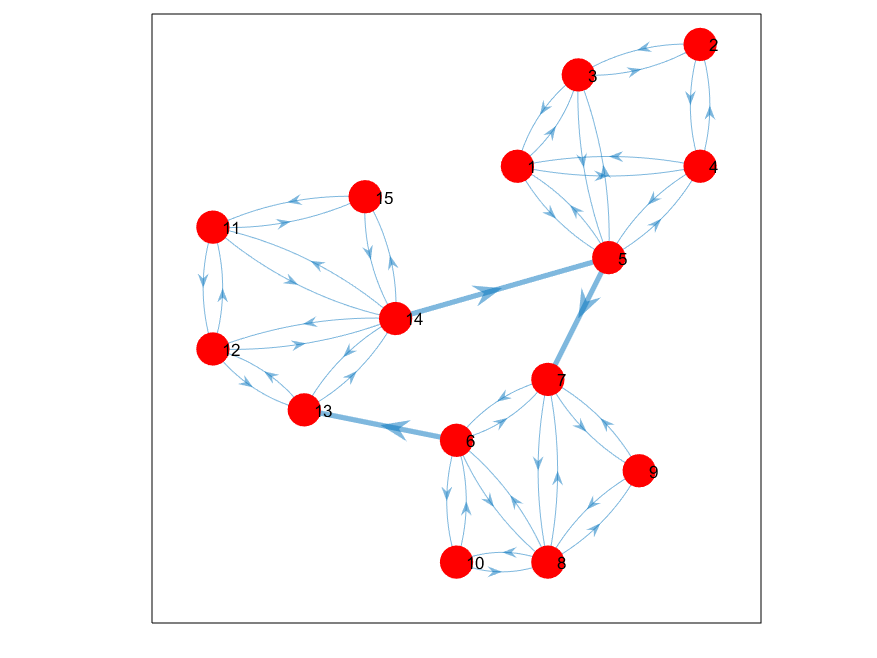}	 
	\caption{Simple graphs with 15 nodes proposed in \citep{Sardellitti17_GFT},
		named here G15a (left), G15b (middle), G15c (right).}
	\label{fig:graph_15}
\end{figure*}

\begin{example} \rm
\label{ex:sarde}
	The graphs shown in Figure \ref{fig:graph_15} were proposed in \citep{Sardellitti17_GFT}
	for illustrating the properties of the eigenvectors of the GFT in a relatively intuitive way.
	Some of them were used also in \citep{Shafipour18_GFT,CCS23}.
	Each graph has three clusters made of five nodes; the edges in a cluster are all digons;
	clusters are connected with directed edges represented with thick arrows;
	the first graph (G15a) has two such directed edges and the second (G15b) has three;
	the third graph (G15c) has also three and they are part of a cycle.
	
	
	\begin{figure*}[h]
	\centering
	\includegraphics[width=0.3\textwidth]{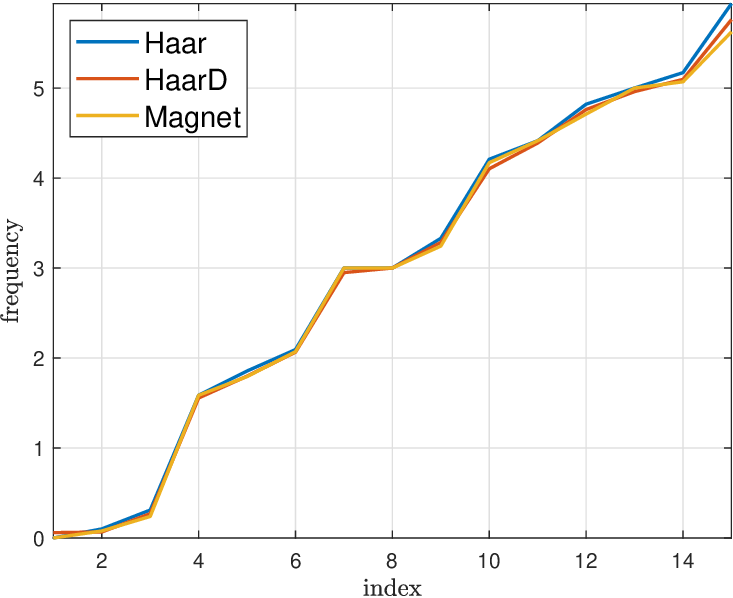}
	\includegraphics[width=0.3\textwidth]{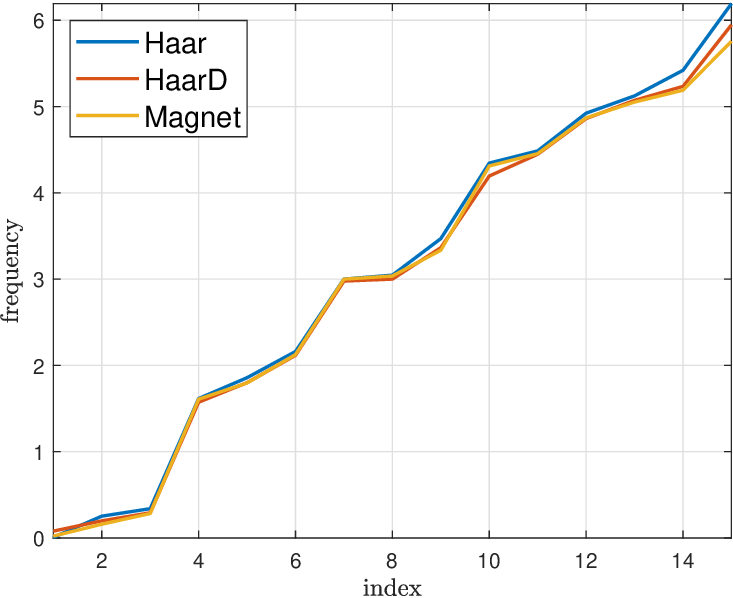}
	\includegraphics[width=0.3\textwidth]{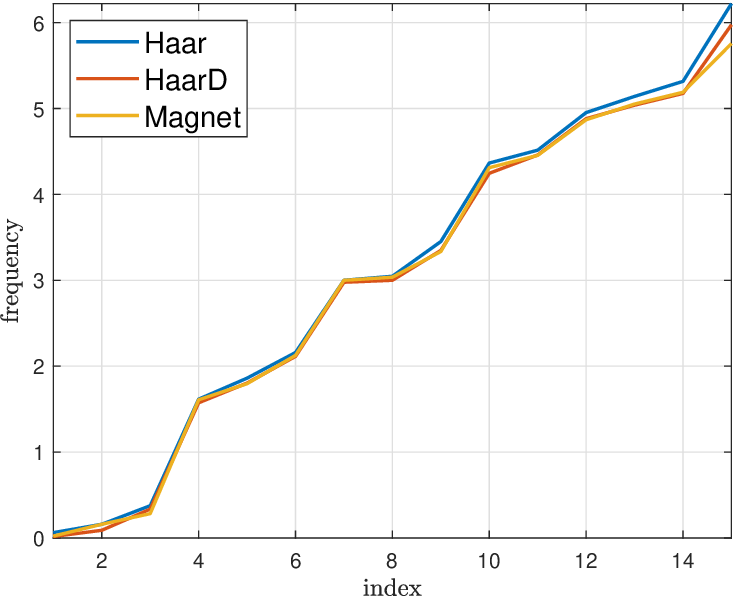} \\	 
	\includegraphics[width=0.3\textwidth]{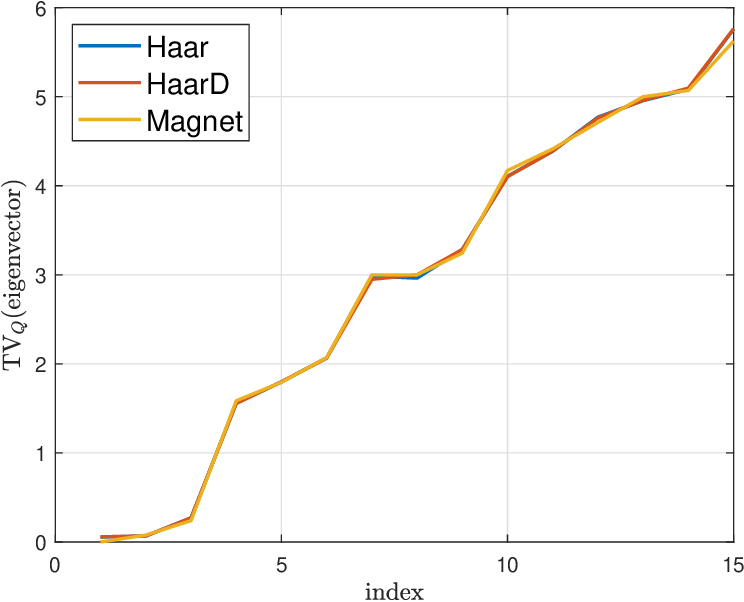}
	\includegraphics[width=0.3\textwidth]{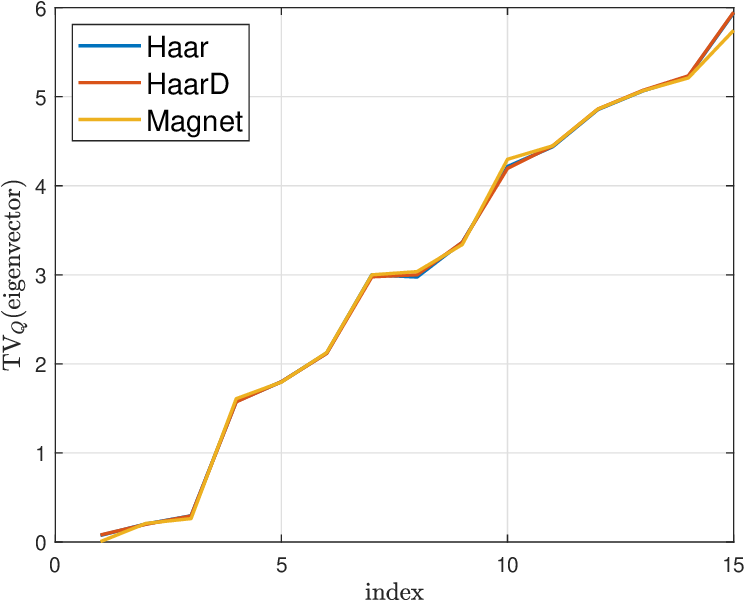}
	\includegraphics[width=0.3\textwidth]{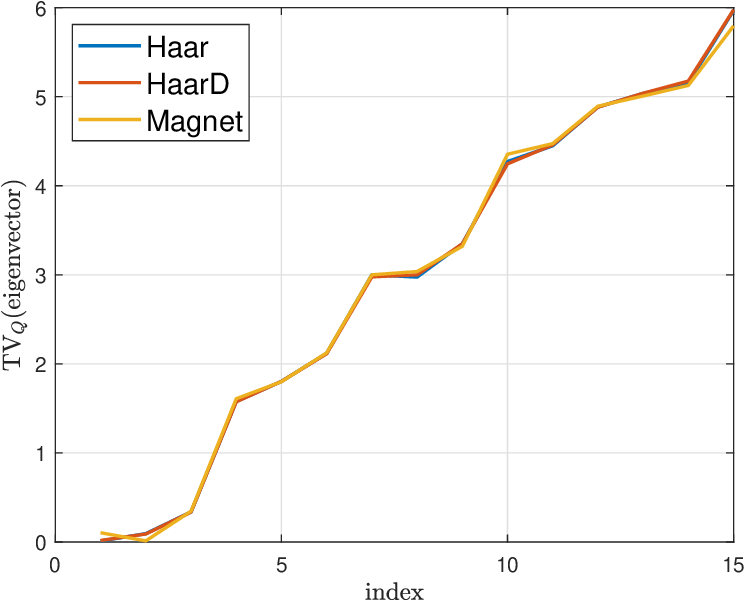} \\	 
	\caption{Top: frequencies associated with the graphs from Fig.\ \ref{fig:graph_15}.
	Bottom: corresponding total variations \eqref{tvQ} of the eigenvectors.}
	\label{fig:graph_15_freq}
	\end{figure*}

	The sorted frequencies associated with the GFTs of the Haar, HaarD, and magnetic (with $q=1/4$) are shown in Figure \ref{fig:graph_15_freq} (top).
	There are insignificant differences between the three sets of frequencies.
	(Note that for $q=1/4$ the magnetic and sign-magnetic Laplacians are again identical.)
	
	Figure \ref{fig:graph_15_freq} (bottom) shows the total variations \eqref{tvQ}
	of the eigenvectors of the same Laplacians.
	The TVs are mostly increasing, as desired, although not always;
	see the first two TV values for the magnetic Laplacian, for G15c.
	The similitude in behavior of the three Laplacians is due mostly
	to the fact that the graphs are not far from symmetry.
	(Remind that for symmetric graphs, the Laplacians are identical.)
	However, the small differences hint to different TV properties.

    The eigenvectors associated with the HaarD-Laplacian for these graphs are illustrated in Appendix B. 
	\QED
\end{example}

\section{Learning with the Haar-Laplacian}
\label{sec:learning}

Throughout this section, we first set the context of the problem that we address by means of learning using the proposed Haar-Laplacian \eqref{haar_lapl}; the fact that it is positive definite is essential. Then, we describe the architecture that we adopted for the graph neural network model.

\subsection{Link prediction problems}

In order to assess the capabilities of the proposed Laplacian on learning tasks, we address the \emph{link prediction} problem, a fundamental task in machine learning on graphs, where the goal is to predict the existence of edges (or relationships) between pairs of nodes.
This is highly relevant in a variety of domains such as social networks, biological networks, and recommender systems, where inferring new or missing connections can provide valuable insights.

Before the advent of GNNs, traditional methods for link prediction relied on graph-based heuristics and feature engineering. 
Some of these methods include approaches based on graph metrics or indices (e.g., common neighbors, Jaccard coefficient, Salton index, Adamic-Adar index, Katz score, etc.).

Traditional heuristics have been complemented or surpassed by modern GNN-based methods, which aim to infer missing links, discover potential new links, or forecast future connections between vertices in an end-to-end manner by learning complex patterns from graph structure and node features.

GNNs excel at capturing the underlying graph topology and feature relationships, thus they prove highly effective for link prediction tasks.
Their advantages include: capturing graph structure by aggregating node and neighbor information (including both local and global relationships in the graph), incorporating node features, and learning node representation which summarize properties of vertices and their neighbors.

We are concerned with three flavors of the link prediction task, which we address in a supervised setting:
\begin{itemize}
	\item \emph{existence} problem - for an ordered pair of vertices, predict whether the directed edge exists or not;
	\item \emph{three-class} problem - for an ordered pair of vertices, predict the existence of the directed edge or of the edge in reversed direction;
	\item \emph{weight prediction} problem - for an ordered pair of vertices, predict the weight of the directed edge (or $0$ in case the edge does not exist).
\end{itemize}

The problem has been addressed in literature by means of both approaches.
For example, in \citep{Ahmad2020missing,Yao2016link} the common neighbor metric is used to predict the existence of links in undirected graphs, but without considering any information about the weights. 
A so-called \textit{reliable-route method} is presented in \citep{Zhao15weight}, which extends unweighted local similarity indices to weighted ones, therefore not only embedding weights in the technique, but also predicting both the existence of links and their weights.
Link weight prediction in directed graphs with arbitrary weights is studied in \citep{Kumar16weight}, where authors introduce the $F \times G$ score, based on two novel measures to characterize a vertex: \textit{fairness} and \textit{goodness}.
A more complex approach is shown in \citep{Qiu20weight}, where decision tree ensembles constructed using such heuristics paired with random forest, gradient boosting decision trees, extreme gradient boosting, and light gradient boosting machine are proposed.
In \citep{Zhang2018link} it is highlighted that each heuristic has a strong assumption on when two vertices are likely to link and their effectiveness on networks where these assumptions fail might be limited. 
Consequently, it is suggested that a more reasonable approach would be to learn a suitable heuristic from a given network instead of using predefined ones. 
So, a heuristic learning paradigm for link prediction has been proposed, aiming to learn a function mapping the subgraph patterns to link existence, thus automatically learning a heuristic suitable for the graph.
Other approaches are based on
inference from neighbor sets \citep{Zhu16weight},
matrix factorization \citep{Chen19weight,Cao22weight},
feature extraction and learning methods \citep{Fu18weight,Liu20new}.

Graph neural networks have also been employed to address the three flavors of the link prediction problem.
Several works tackling the existence and three-class tasks in the context of alternative Laplacians for directed graphs include the magnetic approaches (\citep{Zh21_magnet}, \citep{Fior23sigmanet}) and the quaternionic Laplacian \citep{Fior24quat}. 
Two rather complicated, costly, and limited to undirected graphs techniques are shown in  \citep{zulaika2022wl} and \citep{Liang23weight}. 
These methods require multiple elaborate steps. 
The one from \citep{zulaika2022wl} relies on subgraph extraction, subgraph node ordering through a variation of the Weisfeiler-Lehman algorithm, and -- finally -- training a neural network model with a classical convolutional layer and several fully-connected hidden layers. 
The approach from \citep{Liang23weight} shares the first two steps, then a line graph transformation is additionally required. 
Another difference is that \citep{Liang23weight} employs a GCN model. 
Finally, some recent works are based on graph attention networks \citep{Liu24weight}, \citep{grassia2022weight}.
We note that some of the above approaches are designed for undirected graphs only.

\subsection{Node classification}
We also address the more popular node classification problem, where the objective is to predict the label of individual vertices within a graph using both their intrinsic features and the structure of the graph itself.

There are many approaches to this problem and we mention here only some of the most successful.
Notably, the GNNs based on the magnetic \citep{Zh21_magnet} and sign-magnetic \citep{Fior23sigmanet} Laplacians can be used for node classification as well as for link prediction; they obtain excellent results, surpassing most of their predecessors.
Among these, we mention those using embedding generation via forward propagation (SAGE) \citep{Hamilton17SAGE},
propagation using a personalized PageRank algorithm \citep{Gast19},
convolutional graph networks \citep{Tong20DGCN},
GNN using a new path interpretations for graphs having also negative weights \citep{He22sssnet}.

\subsection{HaarNet architecture}
\label{sec:network_architecture}

We tackle the three sub-tasks of the link prediction and the node classification problems mentioned above by adopting a model similar to the ones used in \citep{Zh21_magnet} and \citep{Fior23sigmanet}.
Spectral graph convolutional layers with the proposed Haar-Laplacian are used to learn node representations based on node features and graph structure. 
Once the node embeddings are learned, a downstream prediction is performed in a manner tailored for the addressed task. Standard PyTorch operations are used, such as {\it Linear}, {\it Dropout}, {\it Conv1D},
and {\it LogSoftmax}.

The core of the employed neural network model is a stack of convolutional layers, with equal dimensions; each of them is followed by a ReLU \citep{relu} nonlinearity.
These layers are constructed as a spectral convolutional filter -- as described in \eqref{eq:filtering}.
To write in a formal manner, let us start by considering a matrix of input graph signals $X \in \C^{N \times C}$, where $C$ denotes the number of input channels, and $\Theta \in \C^{C \times d}$ a matrix of learnable filter parameters of dimension $d$.
The propagation through a convolutional layer of the network can be written as
\begin{equation}
	\label{eq:layer_propagation}
	Y = \sigma ( \Tilde{D}_h^{-\frac{1}{2}} \Tilde{H}_h \Tilde{D}_h^{-\frac{1}{2}} X \Theta ),
\end{equation}
where, following \citep{KiWe16}, $\Tilde{D}_h$ and $\Tilde{H}_h$ (see Definition \ref{def:haarlap}) are computed using $\Tilde{A}_s = A_s + I_N$, and $\sigma$ denotes the nonlinearity.
We apply the nonlinearity in a custom manner in order to accomodate the complex nature of the argument; thus, ReLU is applied individually on the real and imaginary parts of the matrix.

The output $Y \in \C^{N \times d}$ of the convolutions is a matrix of complex node embeddings.

\subsubsection{Link prediction}
Since this problem is focused on edges, we apply an unwind operation, similar to \citep{Zh21_magnet} and \citep{Fior23sigmanet}. 
We concatenate the real and imaginary parts of the learned node embeddings from $Y$ corresponding to source and destination vertices of each edge in the dataset:
\begin{equation}
	\label{eq:unwind}
	Y_{\text{unw}} = \text{\it Unwind} \left( Y \right),
\end{equation}
with $Y_{\text{unw}} \in \R^{e \times 4d}$, where $e$ is the number of edges of the training dataset;
more precisely, the row of $Y_{\text{unw}}$ corresponding to edge $(u,v)$
is $[\text{Re}Y_u \ \text{Re}Y_v \ \text{Im}Y_u \ \text{Im}Y_v]$, where $Y_u$ is the row $u$ of matrix $Y$, i.e., the embedding of node $u$. 
In this manner, we learn a pairwise measure which characterizes the edge between the two vertices.
Then, a dropout layer and a dense (linear) layer are added. 
The linear layer transforms the output to a dimension $s$ matching the sub-task format (1 for the weight prediction, 2 for the existence, and 3 for the three-class).
The result becomes
\begin{equation}
	\label{eq:dropout_linear}
	Z = \text{\it Linear} \left( \text{\it Dropout} \left( Y_{\text{unw}} \right) \right),
\end{equation}
where $Z \in \R^{e \times s}$ is the output of the network.

When addressing the weight prediction sub-task, the set of labels has a size of $e \times 1$, therefore $Z$ is used directly with mean-squared error loss function for training and assessment.
For the existence and three-class flavors of the problem, we also apply logarithmic softmax on the network output $Z$, which, as mention earlier, has a size of $e \times 2$ and $e \times 3$, respectively:
\begin{equation}
	\label{eq:log_softmax}
	Z_{class} = \text{\it LogSoftmax}\left( Z \right),
\end{equation}
where column indices of the maximal values in each row of $Z_{class}$ give the edge prediction label according to the tackled sub-task.
Moreover, the logarithmic softmax further allows for the direct usage of the negative logarithmic likelihood loss.

\subsubsection{Node classification}
After generating the node embeddings $Y \in \C^{N\times d}$ through the convolutional layers, the real and imaginary components are concatenated, thus obtaining $Y_{\text{cat}} \in \R^{N \times 2d}$;
row $u$ of $Y_{\text{cat}}$ is $[\text{Re}Y_u \ \text{Im}Y_u]$.
The concatenated $Y_{\text{cat}}$ is fed into a dropout layer. 
Subsequently, a $1D$ convolution is performed on each embedding, mapping it from the original feature space of size $2d$ to a new space whose dimensionality equals the number of target classes.
The transformation can be written as
\begin{equation}
    \label{eq:dropout_conv1d}
    Z = Conv1D\left( \text{\it Dropout} \left( Y_{\text{cat}} \right) \right),
\end{equation}
where where $Z \in \R^{s \times N}$ is the output of the network, with $s$ denoting the number of classes in this scenario.
Finally, \textit{LogSoftmax} function is applied along the class dimension to produce log-probabilities over the possible labels for each node, as in \eqref{eq:log_softmax}. 
Similarly to the link prediction problem, negative logarithmic likelihood loss is employed.

\section{Numerical results}
\label{sec:results}

\subsection{Learning applications results}

To evaluate\footnote{The programs implementing our approach can be found at \url{https://github.com/theodorbadea/Haar-Laplacian}. The source code itself and the experimental setup heavily rely on the repositories published in \citep{Zh21_magnet} and \citep{Fior23sigmanet}.} our Haar Laplacian on learning tasks, we compare the results with several other alternatives from the literature: magnetic Laplacian (MagNet) \citep{Zh21_magnet}, sign-magnetic Laplacian (SigMaNet) \citep{Fior23sigmanet}, and magnetic signed Laplacian (MSGNN) \citep{He22msgnn}; in the respective works, they are shown to perform better than other methods for learning on directed graphs.
Given the architectures described throughout section \ref{sec:network_architecture}, we train models with various parameter configurations:
\begin{itemize}
	\item number of layers: $2$, $4$, $8$;
	\item layer dimension $d$: $16$, $32$, $64$;
	\item learning rate: $0.001$, $0.005$, $0.01$, $0.05$.
\end{itemize}
So, the resulting number of models is $36$.
We choose a probability of $0.5$ for the dropout layer and a weight decay of $5 \cdot 10^{-4}$ is set to the Adam optimizer.
Moreover, we employ the so-called \textit{renormalization trick} proposed in \citep{KiWe16}. As explained, this procedure alleviates the risk of vanishing or exploding gradients and preserves the identity of the vertex by also aggregating its own information. We ran several experiments without the renormalization and the results are similar.
The learning is supervised. 
So, for link prediction, we split the datasets for training ($80\%$), validation ($5\%$), and testing ($15\%$), as suggested in \citep{Zh21_magnet}. 
The connectivity of the graph is preserved when building each training set by guaranteeing that the graph used for training in each fold contains a spanning tree.
Negative sampling is also used; the number of inexistent edges is taken to be equal to that of actual edges.
For node classification, splits are slightly different: training ($60\%$), validation ($20\%$), and testing ($20\%$).
Experiments are carried out with $k$-cross validation, where $k=10$.

Regarding the training process, we set the maximum number of epochs at $1000$, and enforce an early stopping condition of $200$ iterations without an improvement in the validation results.

Our assessment relies on four real and directed datasets: Telegram \citep{Bovet2020Telegram}; Bitcoin Alpha and Bitcoin OTC \citep{Kumar16weight}; and UC-Social \citep{Opsahl2009UCSocial}.
The first dataset represents a pairwise influence network of Telegram channels, used to analyze interactions concerning the spread of political ideologies. 
It includes $245$ nodes, grouped into four classes, and contains $8912$ edges with positive integer weights up to $7934$.
The Bitcoin datasets originate from cryptocurrency exchanges and are structured as trust networks. In both the Alpha and OTC datasets, users rate each other on a scale from $-10$ to $+10$ (excluding $0$), where $-10$ denotes a scammer and $+10$ indicates a fully legitimate user.
The Bitcoin Alpha dataset comprises $3783$ nodes, with $22650$ positive edges and $1536$ negative edges. In contrast, the slightly larger Bitcoin OTC dataset has $5881$ nodes, with $32029$ positive and $3563$ negative edges, respectively.
UC-Social has $1899$ vertices and $13838$ edges with weights between $1$ and $184$. The graph models the messages sent by users of an online student community from the University of California, Irvine. The nodes and edges represent the users and the messages sent, respectively. The weights highlight the number of messages.

In order to offer an exhaustive perspective, we test on the Bitcoin datasets both with and without the negative weights.
The removal of negative edges is done as in \citep{Fior23sigmanet} and we further denote the resulting graphs with positive edge weights by Bitcoin Alpha+ and Bitcoin OTC+.
As a remark, in the original formulation, MagNet does not support negative weights because the diagonal degree matrix in the original approach \citep{Zh21_magnet} is constructed by summating the columns of $A_s$, which may lead to negative entries.
However, by using absolute values as in \eqref{diag_degree}, the problem disappears; note that otherwise the relations \eqref{magn_lapl}--\eqref{phase_magnet} stand valid for negative weights. 

Normalizing the data would, in many scenarios, support model convergence, numerical stability, bias prevention, and interpretability. 
For MagNet and MSGNN, additionally, a properly chosen normalization is highly beneficial, since it alleviates the phase scaling issues described in Sec. \ref{sec:context_learning}.
We choose to run experiments with a normalization applied to the datasets. In this way, we can fairly compare the results in a setup that does not impair MagNet and MSGNN.
However, given the two different distributions encountered in the weight values, we cannot adopt the same normalization method for all the graphs.
When it comes to the Telegram and UC-Social graphs, a linear scaling operation is not appropriate because the weights range between $1$ and $7934$, most of them noticeably closer to the inferior limit.
As in \citep{Liu20new}, we scale via $w_{new} = e^{-\frac{1}{w_{old}}}$; the new weights are in the interval $\left( 0, 1\right)$.
For the Alpha and OTC graphs, a linear operation that scales the weights to the $\left[-1, 1\right]$ interval is adequate, as it preserves the original distribution and is not affected by, nor it affects the signs. 

For the first two flavors of the link prediction task (existence and 3-class), we report the prediction accuracy,
which is the ratio of correct predictions (in all predictions).
The weight prediction is a regression problem, therefore we report the results in terms of root mean square error (RMSE) and, additionally, $R^2$ metric.
Tables \ref{tab:existence_results}-\ref{tab:weight_r2} show the values obtained for these metrics.

As already mentioned, our experiments emulate an optimal setup for the magnetic Laplacians (MagNet and MSGNN), since weights are normalized and $q = 0.25$.
This ensures a well-behaved phase matrix, with values between $-\pi$ and $\pi$.
The accuracy values shown in Tables \ref{tab:existence_results} and \ref{tab:3class_results} point towards MagNet as the best performing when it comes to existence and 3-class prediction tasks, with an affinity for Bitcoin datasets. Moreover, further analyzing the results for the Bitcoin graphs, a pattern can be observed in Tables \ref{tab:existence_results} and \ref{tab:3class_results}: HaarNet shares the stage with MagNet and MSGNN, whereas SigMaNet is clearly behind. There are two exceptions, however, for the 3-class flavor on Alpha+ and OTC+, where SigMaNet seems to swap places with MSGNN (which displays a considerably lower performance).
Our proposed HaarNet obtains a margin of $0.03$ and $0.11$ over MagNet for the existence and 3-class tasks on Telegram, respectively, while MSGNN and SigMaNet have noticeably lower accuracies. For the UC-Social graph, HaarNet shows a more substantial advantage in the aforementioned evaluation tasks, achieving gains of $0.22$ for existence and $0.32$ over the second best for the 3-class task.
These two problems rely mainly on the directionality of the edges, without a strong dependency on the actual weight assigned to the edges.
On the one hand, the limited range of weights and their variability in the Bitcoin trust networks seem to favor the ability of MagNet and MSGNN to learn the structure of the graph. 
On the other hand, not only does a large proportion of nodes in the Bitcoin graphs have antiparallel edges (digons), most are also with identical weights. Thus, a significant number of entries in the magnetic Laplacian completely overlap with $A_s$.
A similar reasoning can be applied to the 3-class task, the same structure supporting the network in learning the "directed edge exists" predictions.

\begin{table*}[tb]
	\caption{Accuracy ($\%$) for the existence problem}
	\label{tab:existence_results}
	\centering
	\hspace*{-25mm}
	\normalsize
	\begin{tabular}{l|l|l|l|l|l|l} 
		\hline
		Model & Telegram & Alpha & OTC & Alpha+ & OTC+ & UC-Social \\
		\hline
		HaarNet & \textbf{87.09} $\pm$ 0.49 & 87.18 $\pm$ 0.28 & 88.15 $\pm$ 0.28 & 87.61 $\pm$ 0.47 & 88.82 $\pm$ 0.36 & \textbf{85.90} $\pm$ 0.53 \\
		MagNet & 87.06 $\pm$ 0.63 & 87.36 $\pm$ 0.29 & \textbf{88.35} $\pm$ 0.29 & \textbf{87.82} $\pm$ 0.37 & \textbf{89.11} $\pm$ 0.38 & 85.68 $\pm$ 0.59 \\
		SigMaNet & 86.54 $\pm$ 0.46 & 86.66 $\pm$ 0.31 & 87.77 $\pm$ 0.27 & 87.09 $\pm$ 0.37 & 88.43 $\pm$ 0.44 & 85.61 $\pm$ 0.50 \\
		MSGNN & 86.89 $\pm$ 0.55 & \textbf{87.54} $\pm$ 0.40 & 88.25 $\pm$ 0.35 & 87.60 $\pm$ 0.42 & 88.82 $\pm$ 0.30 & 85.13 $\pm$ 0.55 \\
		\hline
	\end{tabular}
\end{table*}

\begin{table*}[htb]
	\caption{Accuracy ($\%$) for the 3-class problem}
	\label{tab:3class_results}
	\centering
	\hspace*{-25mm}
	\normalsize
	\begin{tabular}{l|l|l|l|l|l|l}
		\hline
		Model & Telegram & Alpha & OTC & Alpha+ & OTC+ & UC-Social\\
		\hline
		HaarNet & \textbf{83.13} $\pm$ 0.47 & 84.85 $\pm$ 0.79 & 85.08 $\pm$ 0.39 & 85.10 $\pm$ 0.64 & 85.34 $\pm$ 0.76 & \textbf{77.38} $\pm$ 0.63  \\
		MagNet & 83.02 $\pm$ 0.58 & \textbf{84.99} $\pm$ 0.67 & 84.98 $\pm$ 0.43 & \textbf{85.43} $\pm$ 0.75 & \textbf{85.56} $\pm$ 0.73 & 76.51 $\pm$ 0.80 \\
            SigMaNet & 82.64 $\pm$ 0.67 & 84.46 $\pm$ 0.78 & 84.19 $\pm$ 0.37 & 85.34 $\pm$ 0.48 & 85.28 $\pm$ 0.59 & 77.06 $\pm$ 0.65 \\
            MSGNN & 82.82 $\pm$ 0.76 & 84.70 $\pm$ 0.70 & \textbf{85.11} $\pm$ 0.35 & 84.73 $\pm$ 0.59 & 84.94 $\pm$ 0.59 & 76.21 $\pm$ 0.71 \\
		\hline
	\end{tabular}
\end{table*}

\begin{table*}[htb]
	\tabcolsep=2pt
	\caption{Root Mean Square Error (RMSE) for the weight prediction problem}
	\label{tab:weight_rmse}
	\centering
	\hspace*{-35mm}
	\normalsize
	\begin{tabular}{l|l|l|l|l|l|l}
		\hline
		Model & Telegram & Alpha & OTC & Alpha+ & OTC+ & UC-Social\\
		\hline
		HaarNet & \textbf{0.2533} $\pm$ 0.0039 & \textbf{0.2022} $\pm$ 0.0039 & \textbf{0.2242} $\pm$ 0.0028 & \textbf{0.1380} $\pm$ 0.0033 & \textbf{0.1344} $\pm$ 0.0014 & 0.2430 $\pm$ 0.0031 \\
		MagNet & 0.2534 $\pm$ 0.0039 & 0.2037 $\pm$ 0.0040 & 0.2258 $\pm$ 0.0024 & \textbf{0.1380} $\pm$ 0.0034 & 0.1345 $\pm$ 0.0015 & \textbf{0.2425} $\pm$ 0.0028 \\
        SigMaNet & 0.2578 $\pm$ 0.0035 & 0.2063 $\pm$ 0.0041 & 0.2314 $\pm$ 0.0026 & 0.1389 $\pm$ 0.0032 & 0.1351 $\pm$ 0.0014 & 0.2448 $\pm$ 0.0025  \\
        MSGNN & 0.2565 $\pm$ 0.0040 & 0.2027 $\pm$ 0.0042 & 0.2253 $\pm$ 0.0026 & 0.1382 $\pm$ 0.0036 & 0.1347 $\pm$ 0.0017 & 0.2450 $\pm$ 0.0019 \\
		\hline
	\end{tabular}
\end{table*}

\begin{table*}[t]
	\tabcolsep=2pt
	\caption{Coefficient of determination ($R^2$ score) for the weight prediction problem}
	\label{tab:weight_r2}
	\centering
	\hspace*{-35mm}
	\normalsize
	\begin{tabular}{l|l|l|l|l|l|l}
		\hline
		Model & Telegram & Alpha & OTC & Alpha+ & OTC+ & UC-Social\\
		\hline
		HaarNet & \textbf{0.5314} $\pm$ 0.0128 & \textbf{0.2062} $\pm$ 0.0163 & \textbf{0.2749} $\pm$ 0.0117 & 0.3113 $\pm$ 0.0093 & \textbf{0.3045} $\pm$ 0.0070 & 0.4985 $\pm$ 0.0121 \\
		MagNet & 0.5310 $\pm$ 0.0124 & 0.1943 $\pm$ 0.0172 & 0.2641 $\pm$ 0.0092 & \textbf{0.3118} $\pm$ 0.0109 & 0.3033 $\pm$ 0.0081 & \textbf{0.5004} $\pm$ 0.0107 \\
        SigMaNet & 0.5146 $\pm$ 0.0112 & 0.1737 $\pm$ 0.0183 & 0.2274 $\pm$ 0.0099 & 0.3027 $\pm$ 0.0091 & 0.2972 $\pm$ 0.0060 & 0.4907 $\pm$ 0.0101 \\
        MSGNN & 0.5192 $\pm$ 0.0135 & 0.2021 $\pm$ 0.0208 & 0.2676 $\pm$ 0.0101 & 0.3092 $\pm$ 0.0134 & 0.3007 $\pm$ 0.0114 & 0.4899 $\pm$ 0.0079 \\
		\hline
	\end{tabular}
\end{table*}

We believe that the performance of HaarNet on Telegram and UC-Social is supported by the structure of the datasets, in the sense that Telegram contains an insignificant number of antiparallel edges and UC-Social does not contain any, and, moreover, both have a much larger spectrum of weights.
If for MagNet or MSGNN the phase matrix partly encodes some information about the weights, when it comes to SigMaNet, the phase matrix is entirely dedicated to encoding the directionality, but not complete enough for handling digons.
Although the sign-magnetic Laplacian seems more suitable to sign prediction tasks, nevertheless both our HaarNet and MagNet together with MSGNN perform better in terms of existence problems.

The last series of experiments, addressing the weight prediction flavor of the problem, proves that our Haar Laplacian 
is indeed able to encode all graph information: structure, directionality, and weights.
The task itself closely resembles the existence one; however, the difficulty arises from replacing the binary prediction 
with edge weight values and so shifting the perspective towards the regression family of machine learning problems. 
As can be seen in Tables \ref{tab:weight_rmse} and \ref{tab:weight_r2}, HaarNet outperforms MagNet, SigMaNet, and MSGNN on most graphs and is second to MagNet on Alpha+ and UC-Social by a small difference.
For Telegram and Alpha+, we can see very close performances for HaarNet and MagNet in terms of RMSE and $R^2$ score, with HaarNet winning on Telegram and a tie on Alpha+. 
Slightly different, although HaarNet wins only by $0.0001$ in the RMSE evaluation on OTC+, there is a more significant margin in terms of $R^2$.
We remark especially that the best performance of HaarNet occurs for Alpha and OTC, which are the most difficult problems, since weights can also be negative.
SigMaNet has the worst performance, an outcome that we believe is justified by the fact that this Laplacian is focused on encoding directions rather than weights.
We motivate the superior results of HaarNet by fully preserving the information in the continuous mapping $A \longrightarrow H_h$ and not focusing either on weight or direction, or making a compromise to better encode one to the expense of the other.

\begin{table}[htb]
	\tabcolsep=2pt
	\caption{Node classification results for Telegram}
        \centering
	\label{tab:node_telegram}
	\normalsize
	\begin{tabular}{l|c|c}
		\hline
		Model & L & A \\
		\hline
		HaarNet & 92.34 $\pm$ 2.72 & \textbf{92.77} $\pm$ 3.59 \\
		MagNet & 89.36 $\pm$ 3.00 & 91.49 $\pm$ 3.00 \\
            SigMaNet & 91.28 $\pm$ 4.41 & 89.79 $\pm$ 4.83 \\
            MSGNN & \textbf{92.55} $\pm$ 3.46 & 91.49 $\pm$ 2.52 \\
		\hline
	\end{tabular}
\end{table}

Table \ref{tab:node_telegram} reports node classification results on the Telegram dataset, the only in our pool of graphs naturally suited for such task. We include two evaluation metrics:
\begin{itemize}
    \item L: accuracy reported for the model with the lowest value of the loss function on the validation set;
    \item A: accuracy reported for the model with the greatest value of the accuracy on the validation set.
\end{itemize}
Our HaarNet achieves strong performance in both settings, obtaining $92.34$ for L (second best) and the highest result of $92.77$ for A, indicating its robustness and consistency in these evaluation metrics.
MSGNN slightly outperforms HaarNet in the L setting with a slightly higher score of $92.55$, but falls short in A. This suggests its performance may be less stable depending on the task variation.
MagNet and SigMaNet trail behind both HaarNet and MSGNN, particularly in the L setting, where their accuracies drop below $91$.
Overall, HaarNet emerges as the most balanced model, offering high and consistent accuracy in both metrics, while MSGNN excels only in one.
 
\subsection{Signal processing results}

We give here some results in denoising on graphs, a standard graph signal processing problem.
We use both the Haar \eqref{haar_lapl} and HaarD \eqref{haarlap_d} Laplacians, 
having in mind that the latter has a better frequency domain interpretation.
Our purpose is to compare the Laplacians for directed graphs on a standard problem.

\begin{figure}[t]
	\centering
	\includegraphics[width=0.45\textwidth]{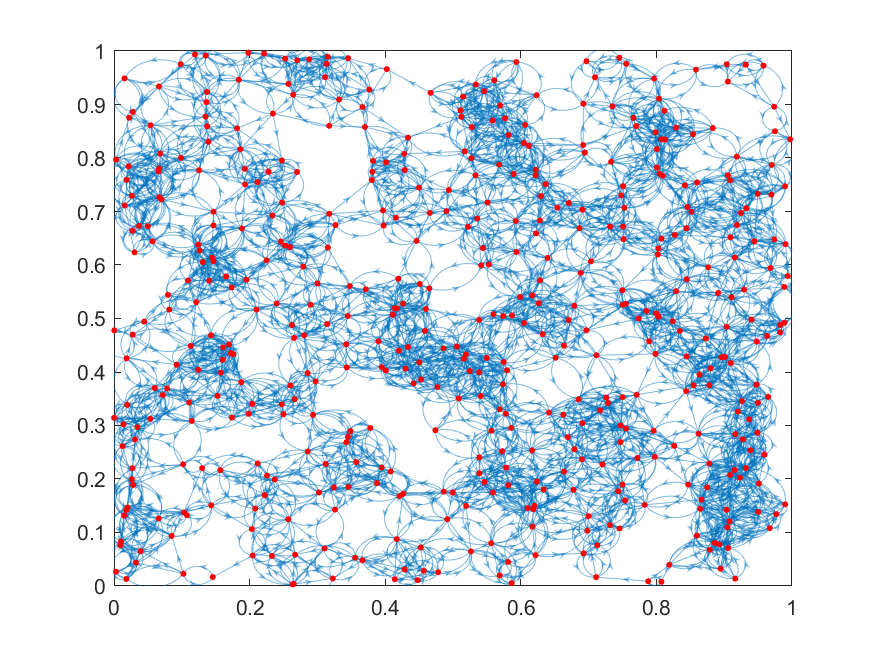}	 
	\caption{Directed graph based on geometric distance, with $N=500$.}
	\label{fig:graph_geom}
\end{figure}

We build a random geometric graph as follows.
The $n$ nodes are given uniformly random coordinates in the square $[0,1] \times [0,1]$.
Each node $u$ is connected with the nodes $v$ that are at distance at most $r$ from it;
the direction of the edge is random, with equal probability of $(u,v)$ and $(v,u)$.
With probability $p$, there are edges in both directions: $(u,v), (v,u) \in {\cal E}$.
The weights of the edges are uniformly random in the interval $[w_{\min}, w_{\max}]$.
Figure \ref{fig:graph_geom} shows an example of such graph, with $N=500$, $r=2/\sqrt{n}$,
and $p=0.5$.

We consider a graph signal defined by $z(u) = 10 + 10x(u) + 5y(u)$,
where $x(u)$ and $y(u)$ are the coordinates of node $u$;
the choice of a plane was made to ensure that the signal is smooth;
we have obtained similar results with other simple signals.
To the signal, we add Gaussian noise with standard deviation $\sigma$;
let $\check z$ be the noisy signal.
The weights of the graph are generated in the interval defined by
$w_{\min} = 0.8$, $w_{\max} = 1.2$.
The setup of the experiment is similar to that from \citep{CCS23}, but on a larger graph.

Denoising is performed with the GFT in a standard way, by applying
a lowpass filter in the frequency domain.
More precisely, we use a simple rectangular window.
The first $m$ coefficients of the frequency domain signal GFT$(\check z) = U^H \check z$,
where $U$ is the matrix of eigenvectors, see \eqref{eigLh}, are
preserved and the remaining coefficients are set to zero.
Then, the inverse GFT is applied, thus obtaining the denoised signal $\hat z$.
The SNR is
\[
\text{SNR} = - 20 \log_{10} \frac{\|\hat z - z\|}{\|z\|}.
\]

\begin{figure}[t]
	\centering
	\bt{cc}
	\includegraphics[width=0.225\textwidth]{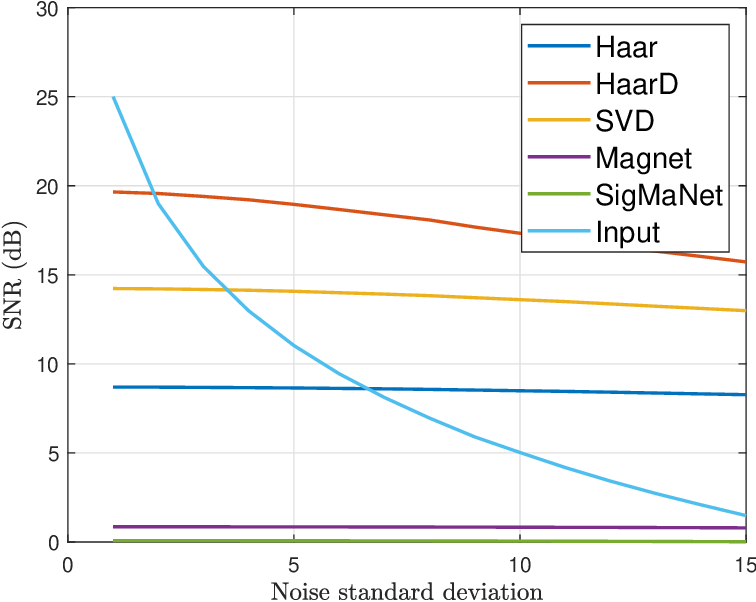} & 
	\includegraphics[width=0.225\textwidth]{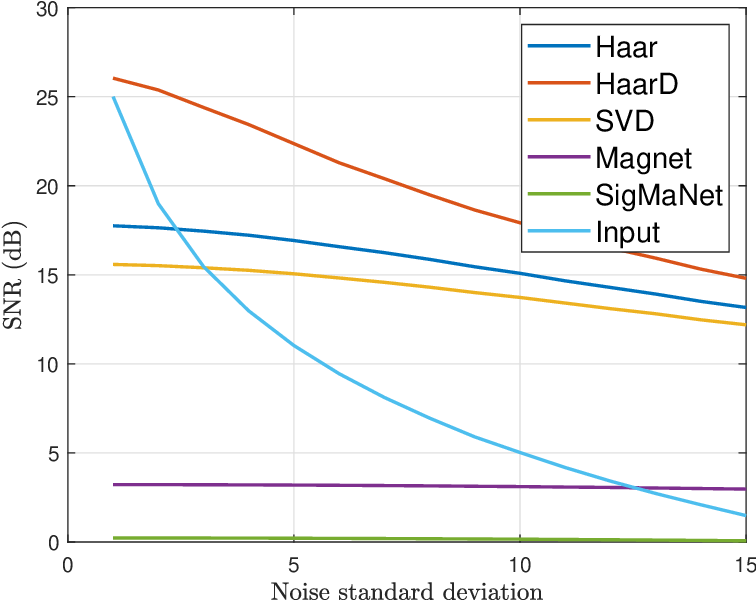} \\
	\includegraphics[width=0.225\textwidth]{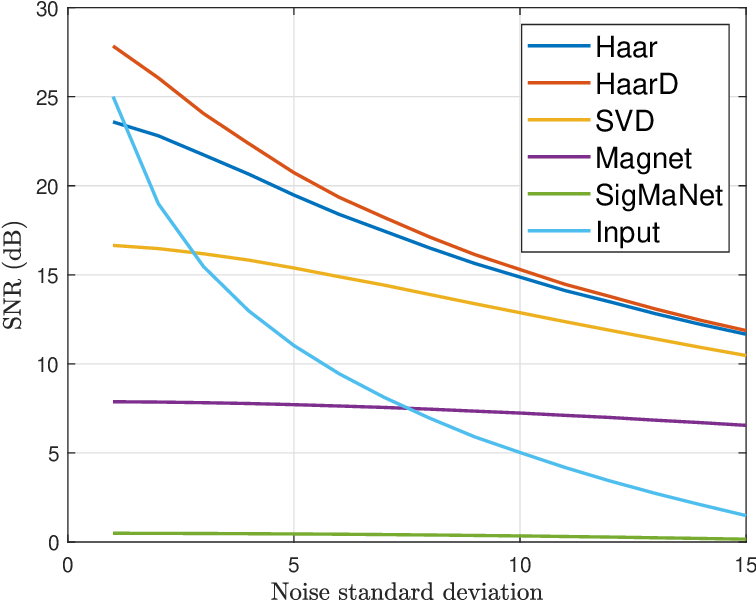} & 
	\includegraphics[width=0.225\textwidth]{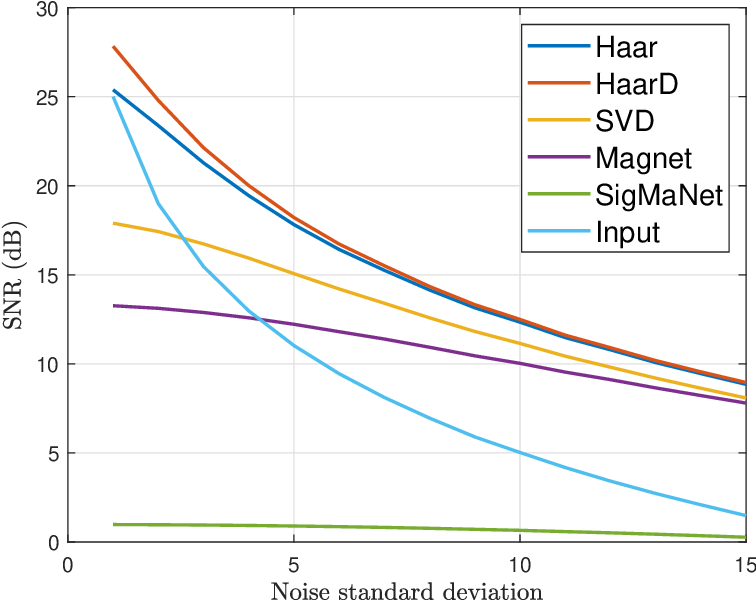}
	\et
	\caption{SNRs obtained by GFT truncation to $m=10$ (up, left), $m=25$ (up, right),
		$m=50$ (down, left), and $m=100$ (down, right), as a function of the noise standard deviation.
		Graph parameters: $N=500$, $p=0.5$, $w_{\min} = 0.8$, $w_{\max} = 1.2$.}
	\label{fig:denoise_1}
\end{figure}

Figure \ref{fig:denoise_1} shows the denoising results for five GFTs:
our Haar- and HaarD-Laplacians, the SVD based one \citep{CCS23},
and the magnetic (with $q=1/4$) \citep{Zh21_magnet}
and sign magnetic \citep{Fior23sigmanet} Laplacians.
The SNRs are averaged over 10 different graphs and 100 runs for each graph 
(with different noise realizations).
One can see that the HaarD-Laplacian gives the best result in all cases.
For a small number $m$ of preserved coefficients, the SVD is better than the
Haar-Laplacian; however, the situation is reversed when $m$ grows.
The magnetic Laplacian gives rather poor results, but is able to improve performance
with growing $m$.
The sign-magnetic Laplacian, which is an artificial construction meant mostly for
coding the properties of the graph rather than for offering a direct interpretation,
is obviously useless for denoising.

\begin{figure}[t]
	\centering
	\includegraphics[width=0.45\textwidth]{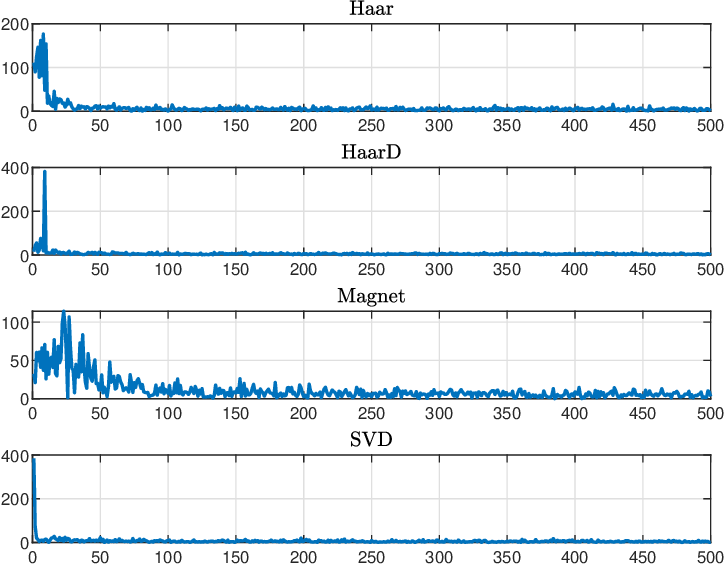}
	\caption{Amplitude of GFT coefficients for a single signal realization, for Haar,
		HaarD, magnetic, and SVD Laplacians (from top to bottom).
		Graph parameters: $N=500$, $p=0.5$, $w_{\min} = 0.8$, $w_{\max} = 1.2$.}
	\label{fig:gft_coef}
\end{figure}

Figure \ref{fig:gft_coef} shows the amplitudes of the GFT coefficients of
a signal realization, on the first graph of the above simulation.
The standard deviation of the noise is $\sigma=5$, which corresponds to
an input SNR of $11.1$~dB.
We see that the SVD is able to compress the most of the signal in the first
few (low frequency) coefficients;
the HaarD GFT needs more coefficients and the Haar GFT even more.
However, at high frequencies, the Haar and HaarD Laplacians have smaller coefficients;
the average amplitude of the last 400 coefficients is $4.93$ and $4.91$,
respectively.
For SVD, the corresponding value is $5.13$.
So, overall, the reconstruction with $m=10$ coefficients using HaarD is
more accurate (SNR $=24.1$~dB) than that using SVD (SNR $=14.3$~dB)
or Haar (SNR $=12.6$~dB).
For larger $m$, Haar can beat SVD, as seen in average in Figure \ref{fig:denoise_1};
for the current realization, this happens already for $m=20$.
The coefficients of the magnetic GFT are spread over a large interval of frequencies
and hence denoising succeeds only for large $m$; the average amplitude of the last 400 coefficients is $6.70$.

\begin{figure}[t]
	\centering
	\bt{cc}
	\includegraphics[width=0.225\textwidth]{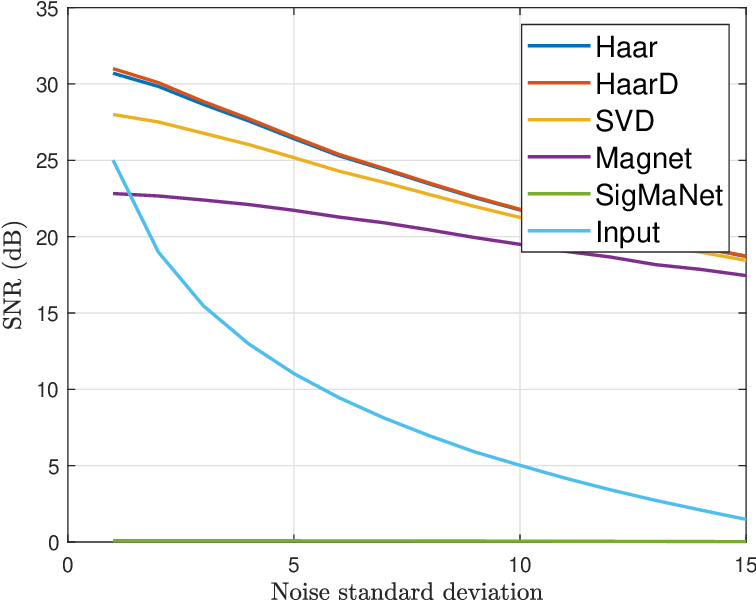} & 
	\includegraphics[width=0.225\textwidth]{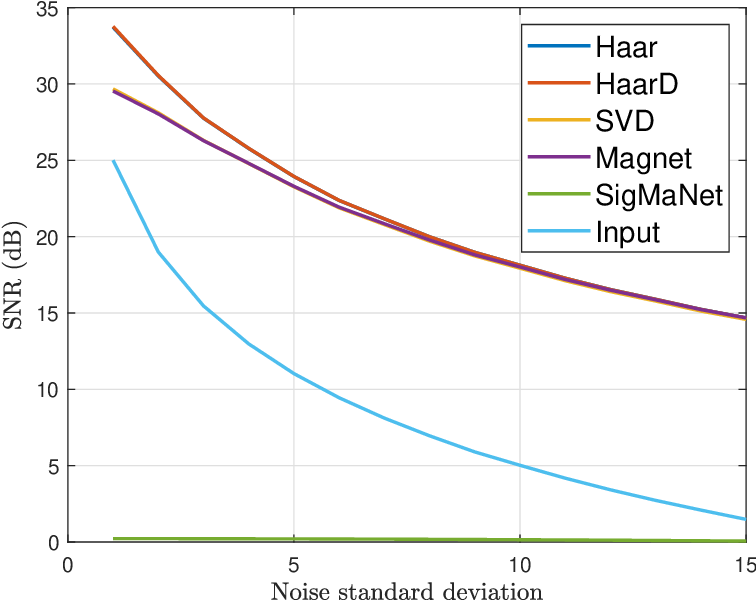} \\
	\includegraphics[width=0.225\textwidth]{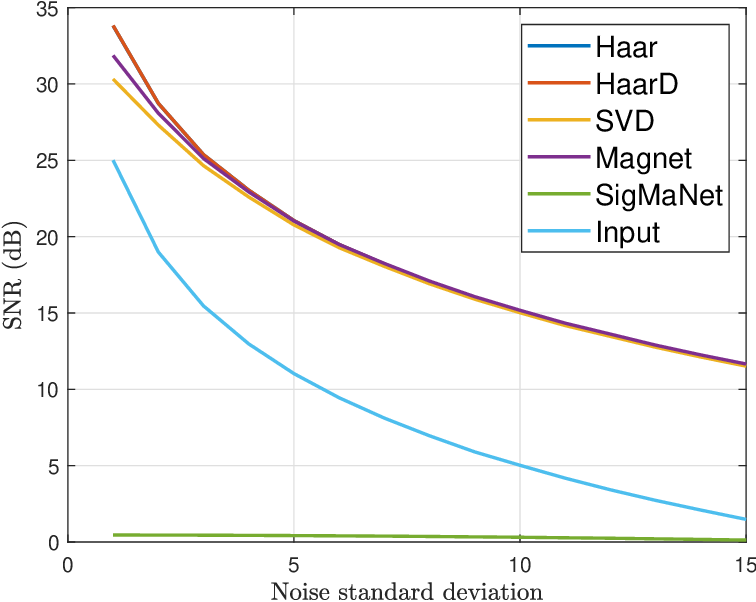} & 
	\includegraphics[width=0.225\textwidth]{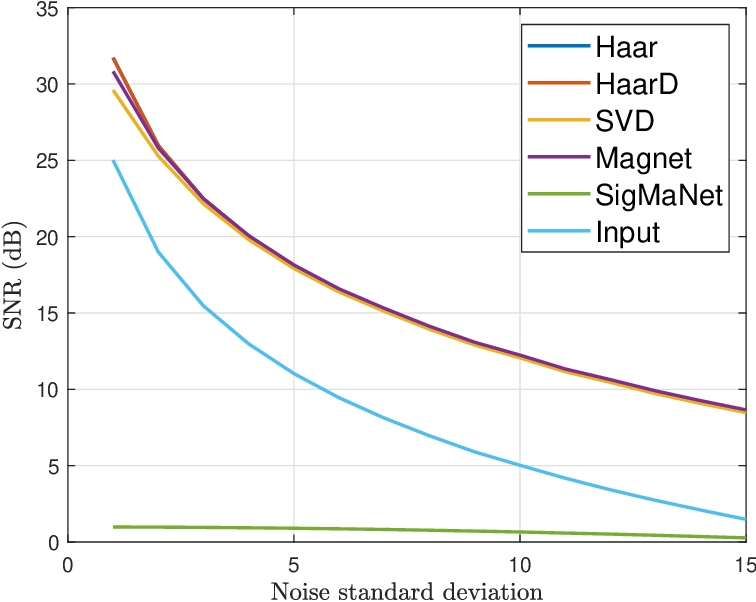}
	\et
	\caption{SNRs obtained by GFT truncation to $m=10$ (up, left), $m=25$ (up, right),
		$m=50$ (down, left), and $m=100$ (down, right), as a function of the noise standard deviation.
		Graph parameters: $N=500$, $p=1$, $w_{\min} = 0.8$, $w_{\max} = 1.2$.}
	\label{fig:denoise_2}
\end{figure}

Figure \ref{fig:denoise_2} displays the denoising results for $p=1$, the other parameters being
the same as above.
This means that all nodes that are connected have edges with both orientations between them.
Now HaarD is slightly better than Haar only at high SNR, otherwise their results
being practically the same.
They are both marginally better than the SVD and better than the magnetic Laplacian.
From both figures it is also visible that, as $\sigma$ grows, the best SNR is obtained
for smaller values of $m$.

\section{Conclusions}
\label{sec:concl}

In this paper, we have proposed a novel Laplacian matrix, called Haar-Laplacian,
and a variation called HaarD-Laplacian, using a different diagonal.
They are designed to enable spectral techniques for directed graphs. Our approach enjoys several key properties that distinguish it from existing methods, including continuity, scaling robustness, sensitivity, and the inherent consideration of directionality in the structure of the graph. The theoretical foundations of our method have been rigorously established, ensuring that the proposed Laplacian is in conformity with the spectral graph theory.

Then, we have addressed both learning and signal processing tasks and we have shown that it is a viable and promising tool for a wide range of graph-based applications.

\section*{Acknowledgments}

The authors thank the reviewers for their thorough and constructive comments,
which helped improve the quality of this paper.




\section*{Appendix A. Proof of Prop. \ref{p:haarD}}
\label{app:proof}
	
We denote $z=\xi + i \eta$, with $\xi, \eta \in \R^n$.
With
$$Q=\baa{cc} 1 & -1-i \\ -1+i & 1 \eaa,$$
the distance \eqref{distQ} is
\[
d(z_u,z_v) = |z_u|^2 + |z_v|^2 - 2(\xi_u \xi_v + \eta_u \eta_v - \xi_u \eta_v + \xi_v \eta_u).
\]
Note that the symmetry property $d(z_u,z_v) = d(z_v,z_u)$ is not satisfied,
hence $d$ is not properly a distance; similarly, the directed variation defined
in \citep{Sardellitti17_GFT,Shafipour18_GFT} is based on an asymmetric measure.
Inserting the above in \eqref{tvQ} gives
\be
\ba{rcl}
\text{TV}_Q(z) & = & \left| \frac{1}{2} \sum_{u,v} a_{uv} (|z_u|^2 + |z_v|^2) \right. \\
& & - \left. \ \sum_{u,v} a_{uv}( \xi_u \xi_v + \eta_u \eta_v - \xi_u \eta_v + \xi_v \eta_u) \right|
\label{tvQ_app}
\ea
\ee
The first sum in \eqref{tvQ_app} can be written
\[
\sum_{u,v} \frac{a_{uv}+a_{vu}}{2} |z_u|^2 = z^H D_s z,
\]
with $D_s$ in \eqref{Ds}.
The second sum in \eqref{tvQ_app} is simply $z^H H_h z$, with $H_h$ in \eqref{haarlap_adj}.
Hence, in view of \eqref{haarlap_d}, the equality \eqref{haard_tv} is satisfied.

\section*{Appendix B. Eigenvectors of the Haar-Laplacian GFT}
\label{app:vectors}

\begin{figure*}[p!]
		\centering
        \bt{ccc}
		\includegraphics[width=0.3\textwidth]{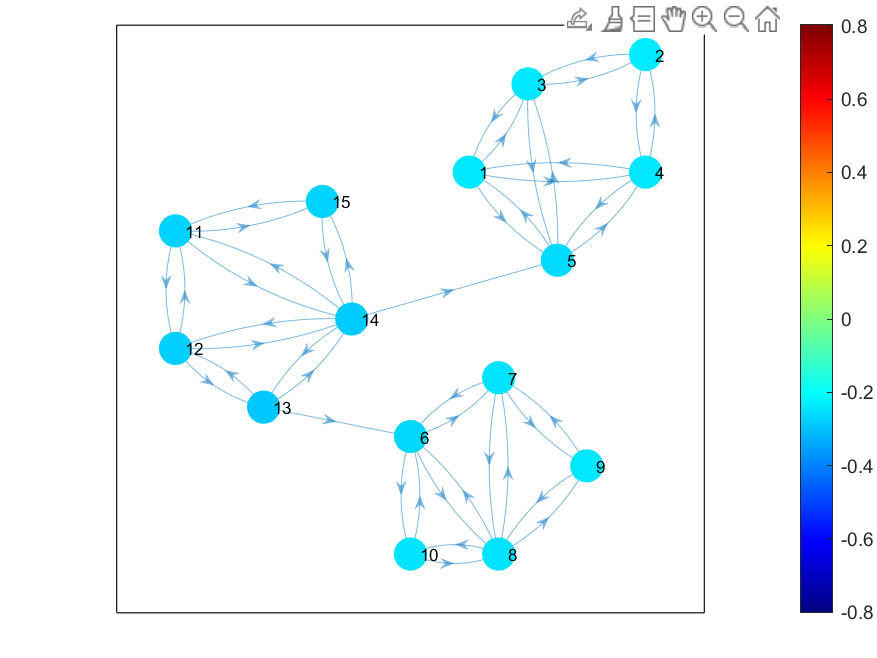} & 
		\includegraphics[width=0.3\textwidth]{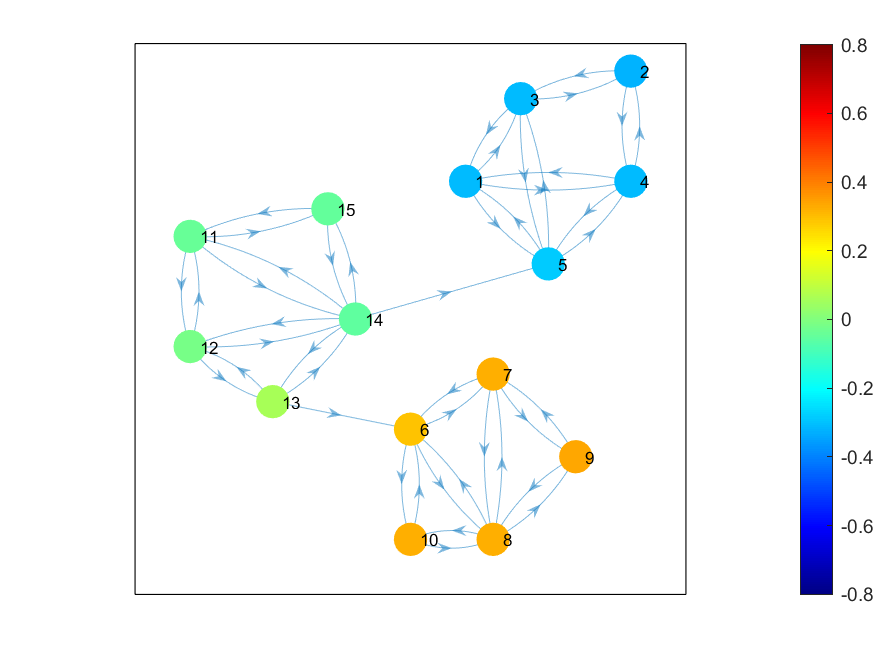} &
		\includegraphics[width=0.3\textwidth]{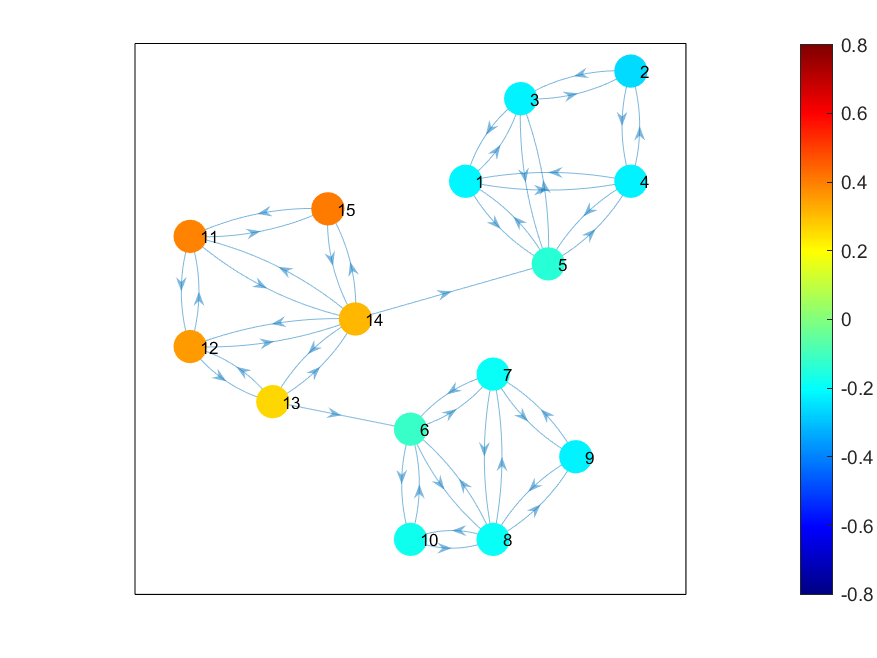} \\
		\includegraphics[width=0.3\textwidth]{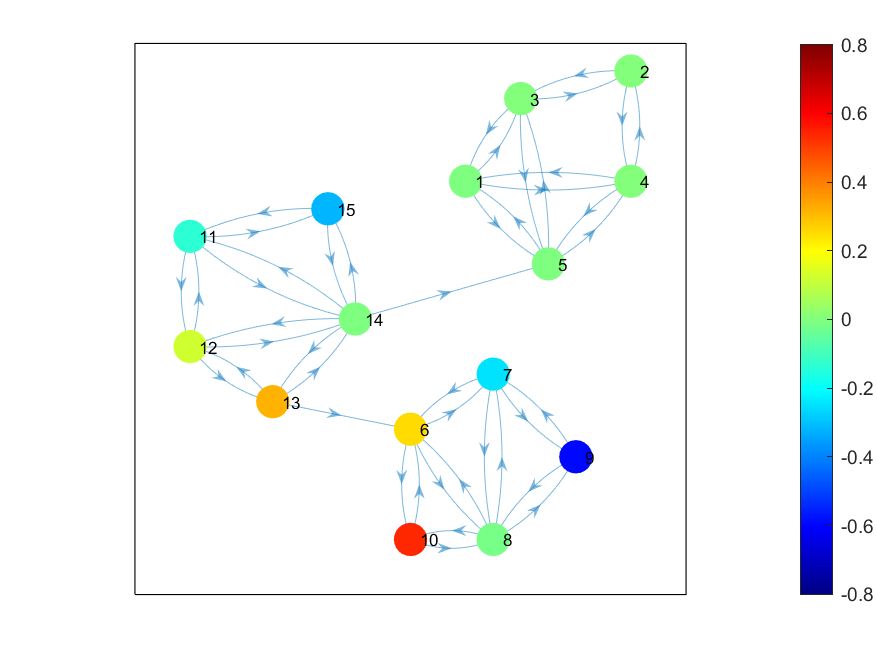} & 
		\includegraphics[width=0.3\textwidth]{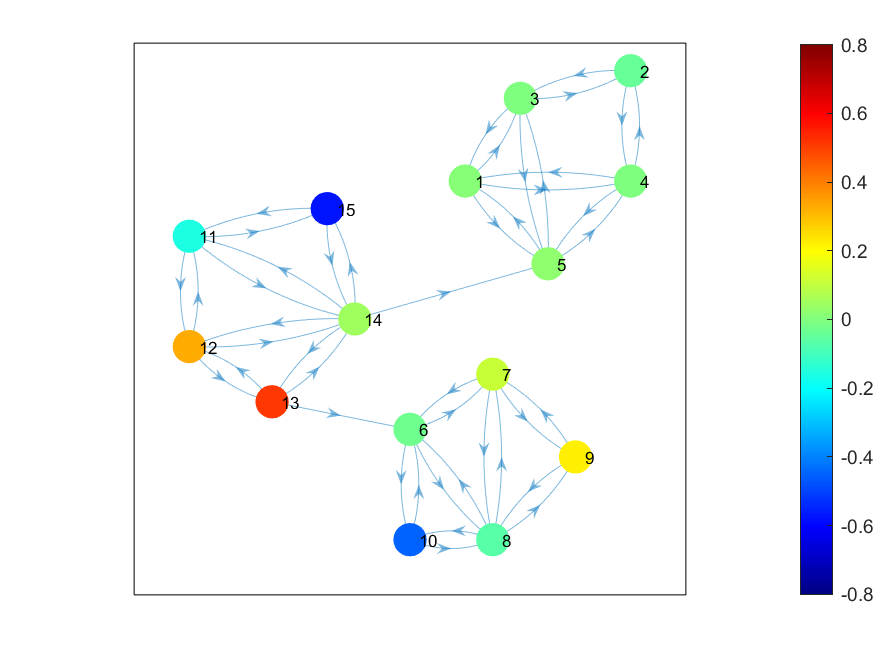} &
		\includegraphics[width=0.3\textwidth]{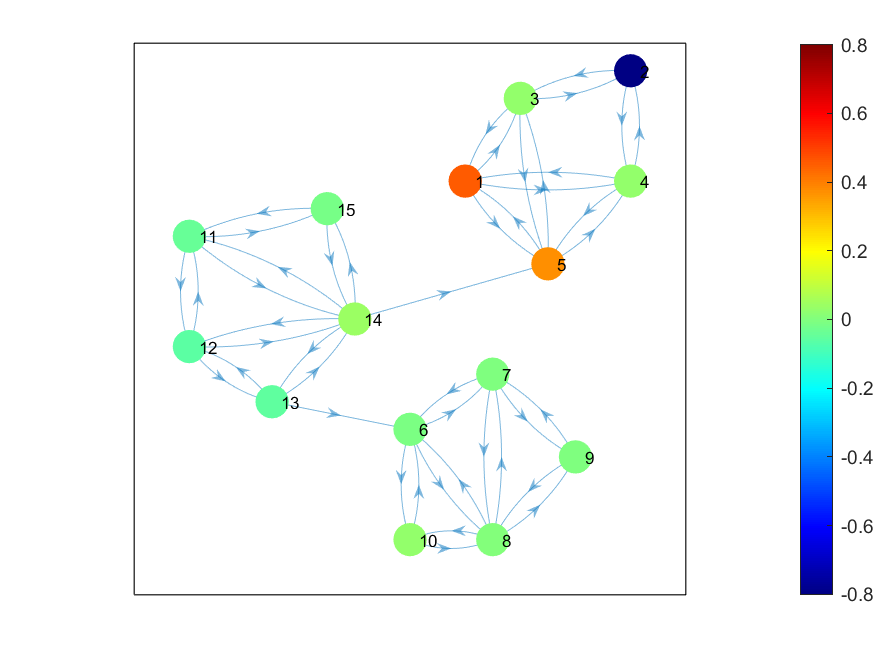} \\
		\includegraphics[width=0.3\textwidth]{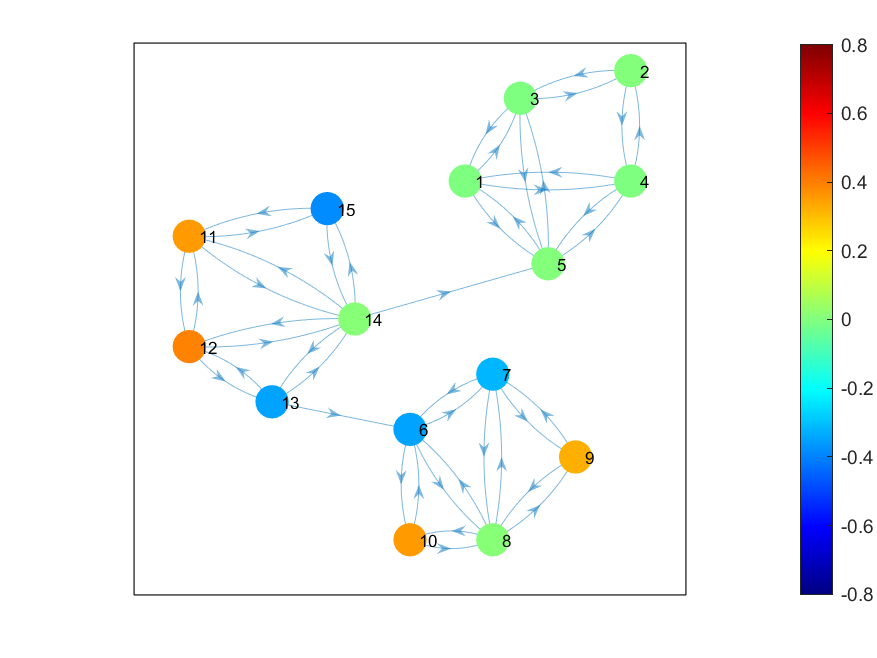} & 
		\includegraphics[width=0.3\textwidth]{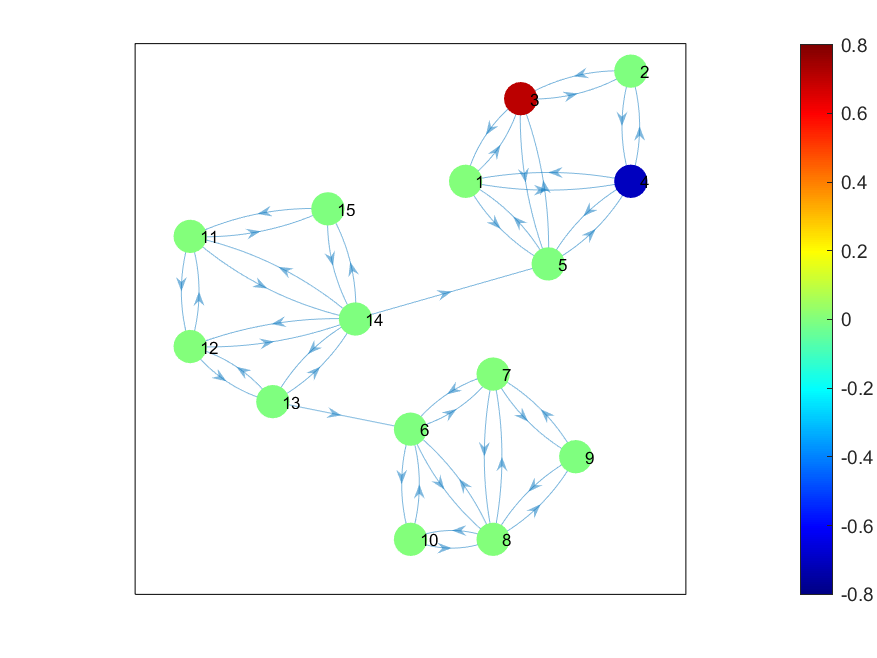} &
		\includegraphics[width=0.3\textwidth]{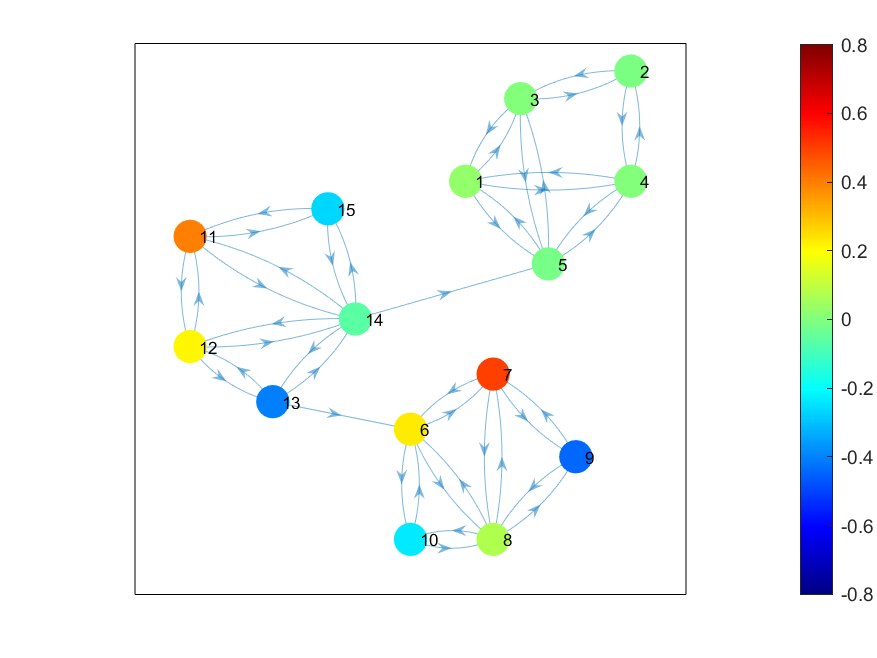} \\
		\includegraphics[width=0.3\textwidth]{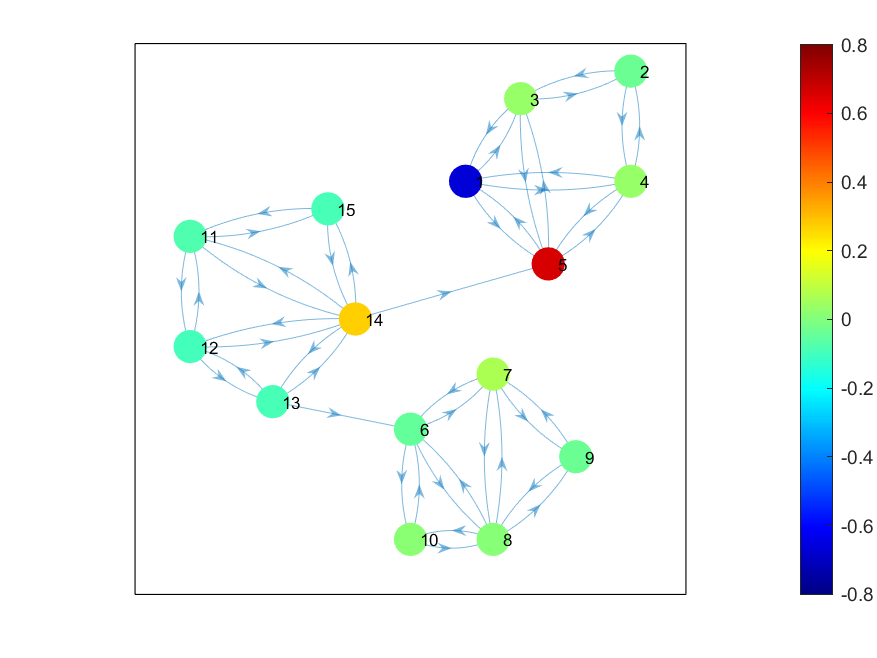} & 
		\includegraphics[width=0.3\textwidth]{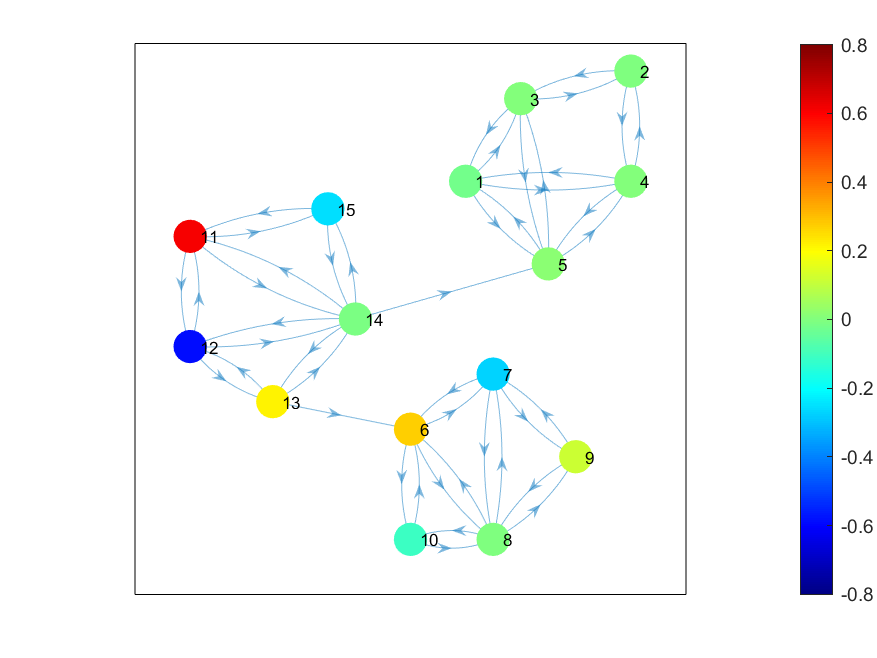} &
		\includegraphics[width=0.3\textwidth]{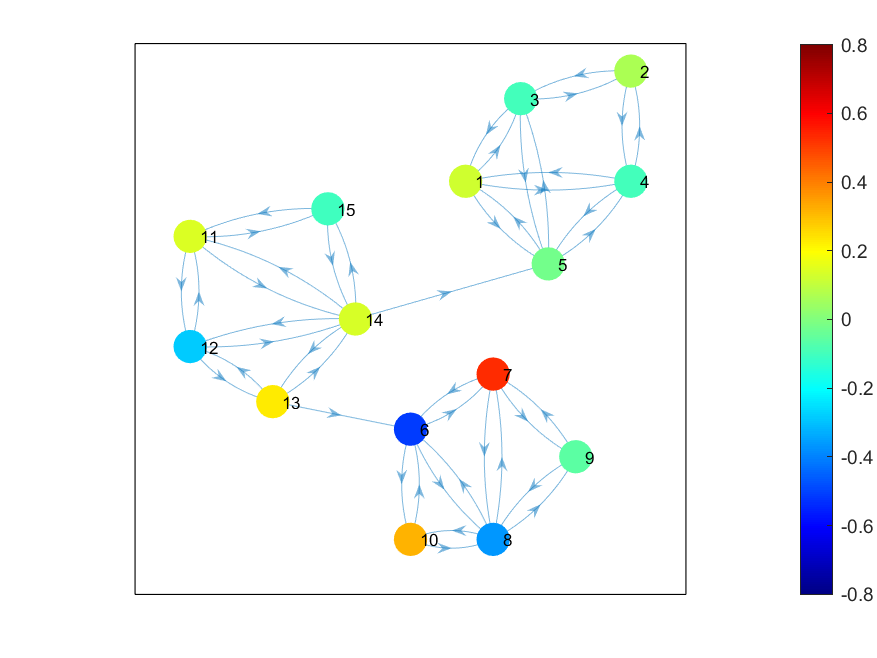} \\
		\includegraphics[width=0.3\textwidth]{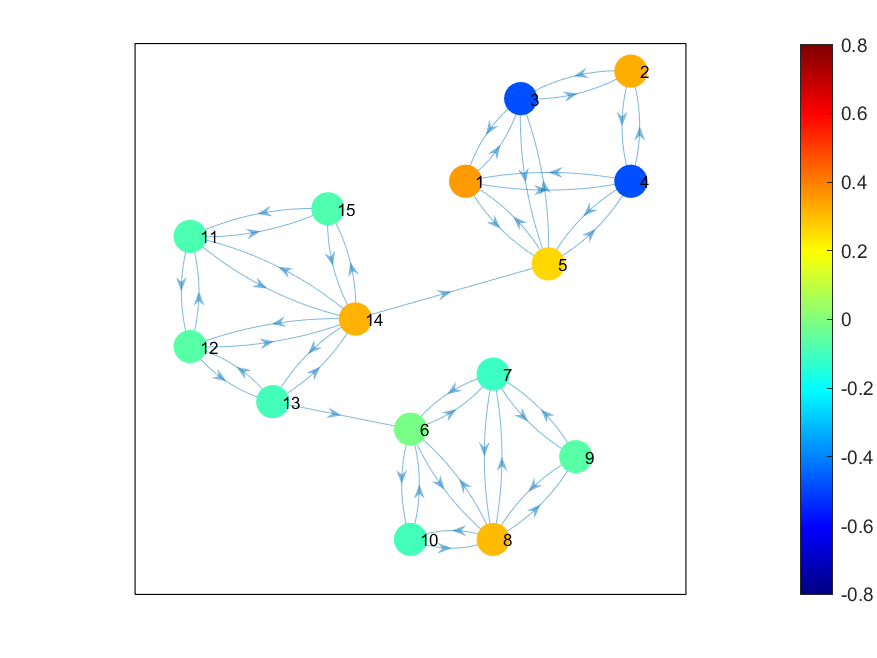} & 
		\includegraphics[width=0.3\textwidth]{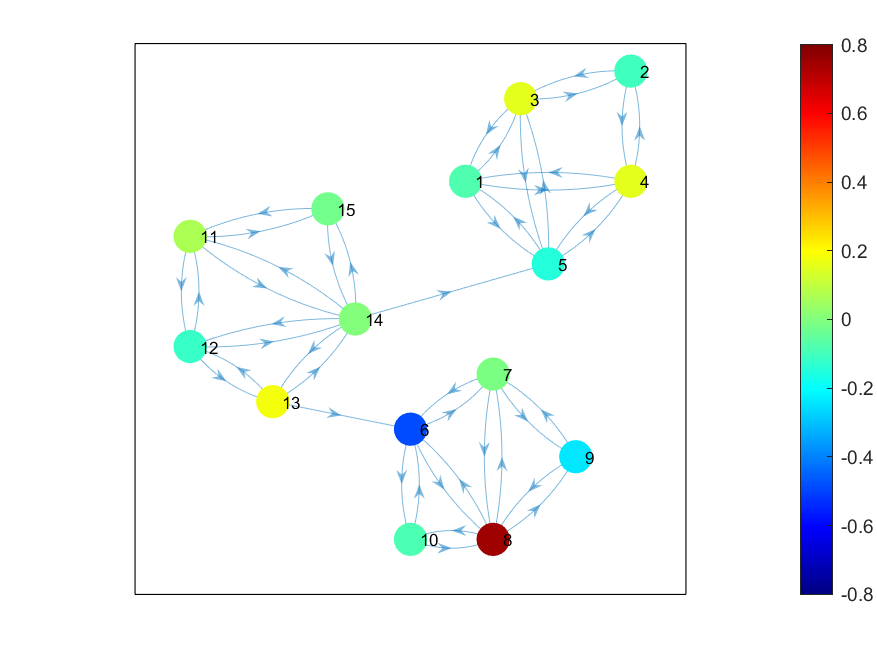} &
		\includegraphics[width=0.3\textwidth]{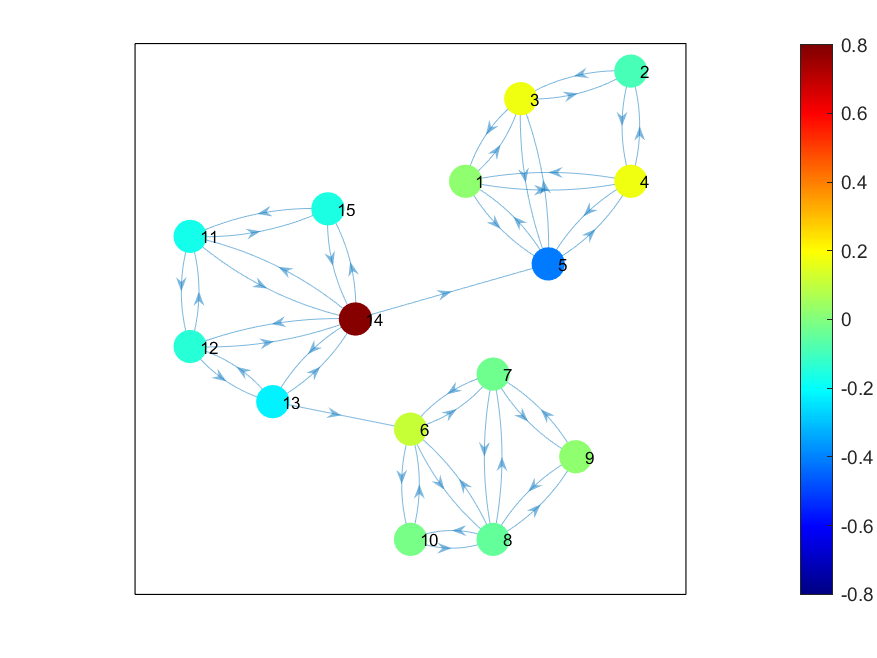}
		\et
		\caption{Eigenvectors of graph G15a.}
		\label{fig:eigv_15_a}
	\end{figure*}

\begin{figure*}[p!]
		\centering
		\bt{ccc}
		\includegraphics[width=0.3\textwidth]{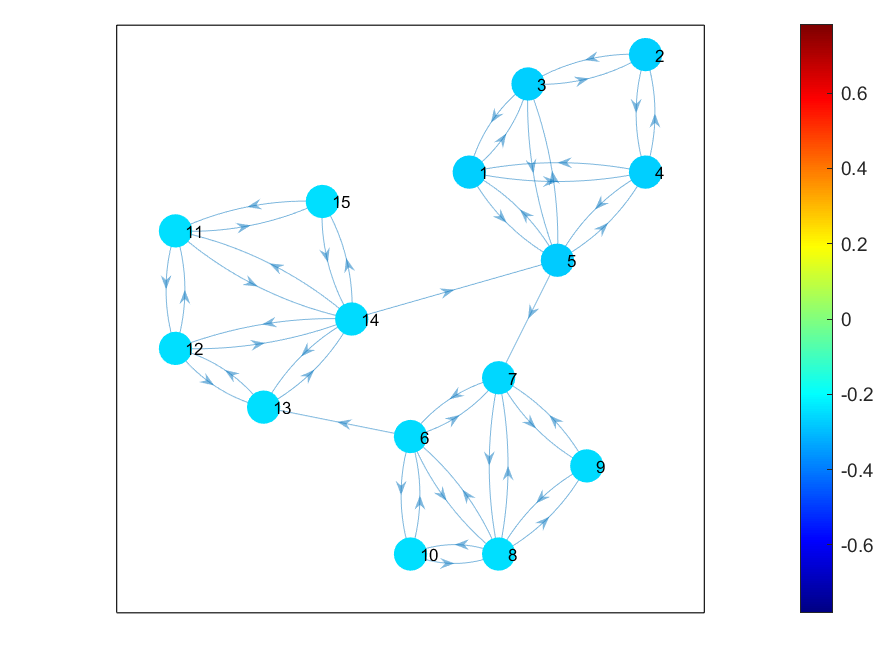} & 
		\includegraphics[width=0.3\textwidth]{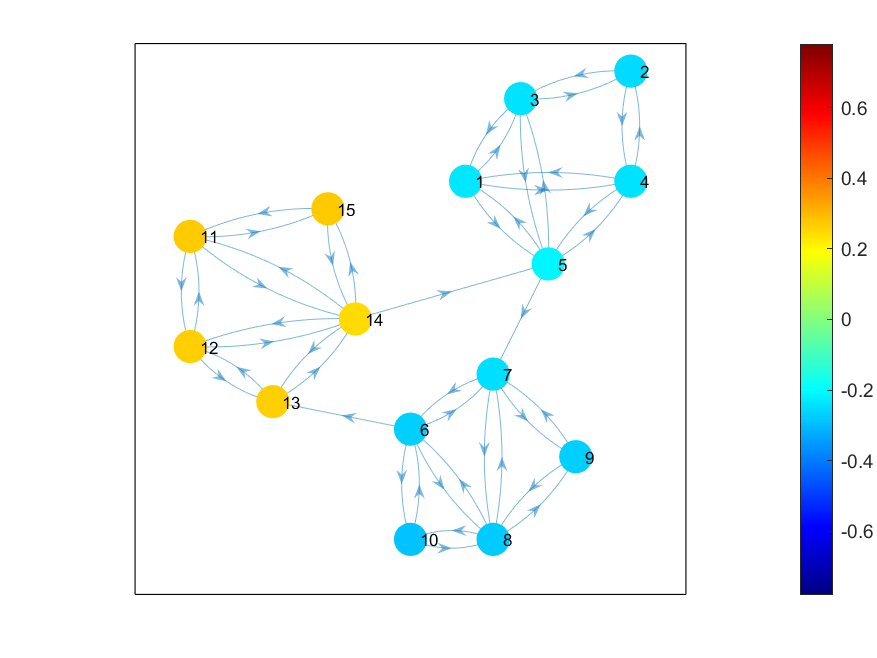} &
		\includegraphics[width=0.3\textwidth]{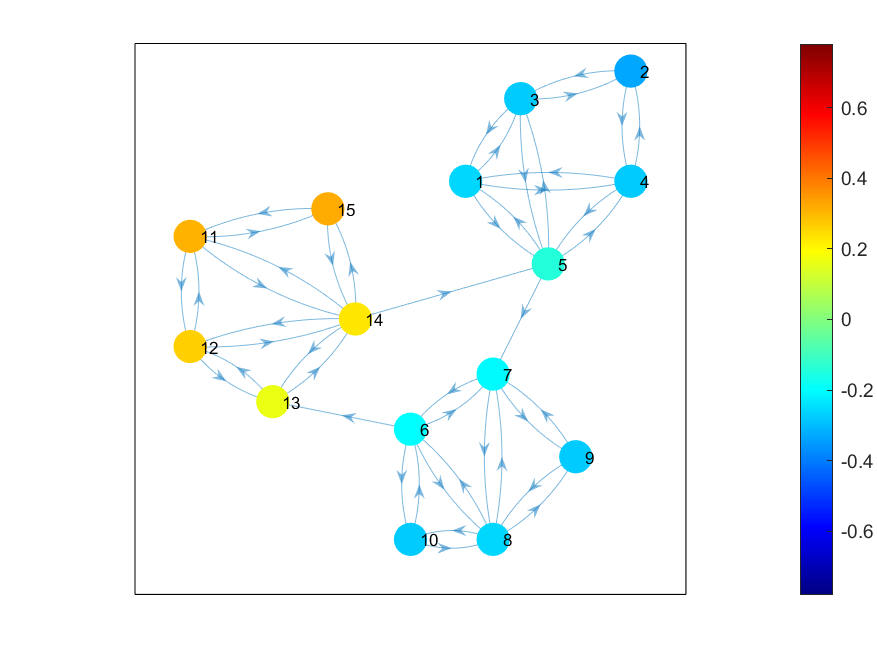} \\
		\includegraphics[width=0.3\textwidth]{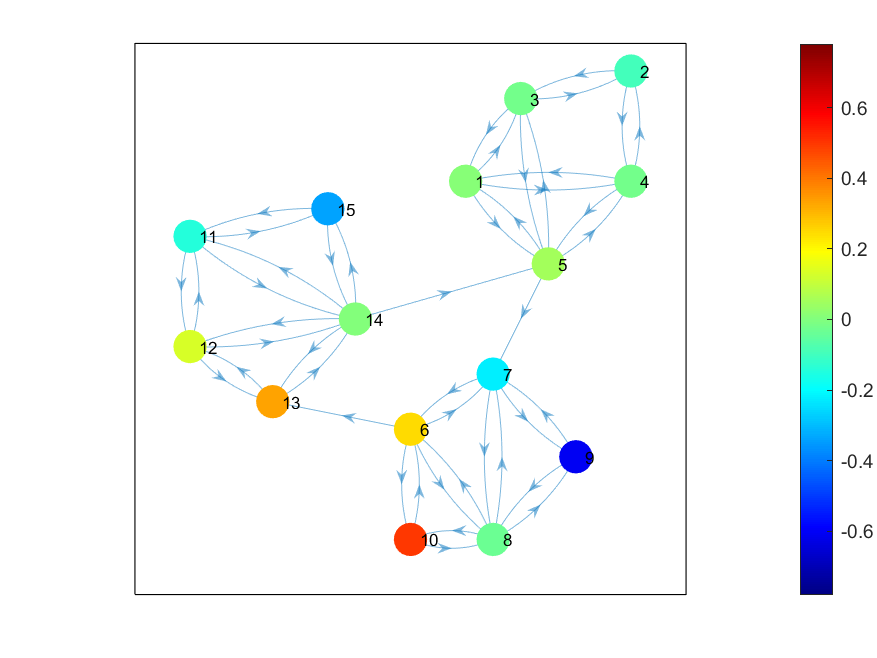} & 
		\includegraphics[width=0.3\textwidth]{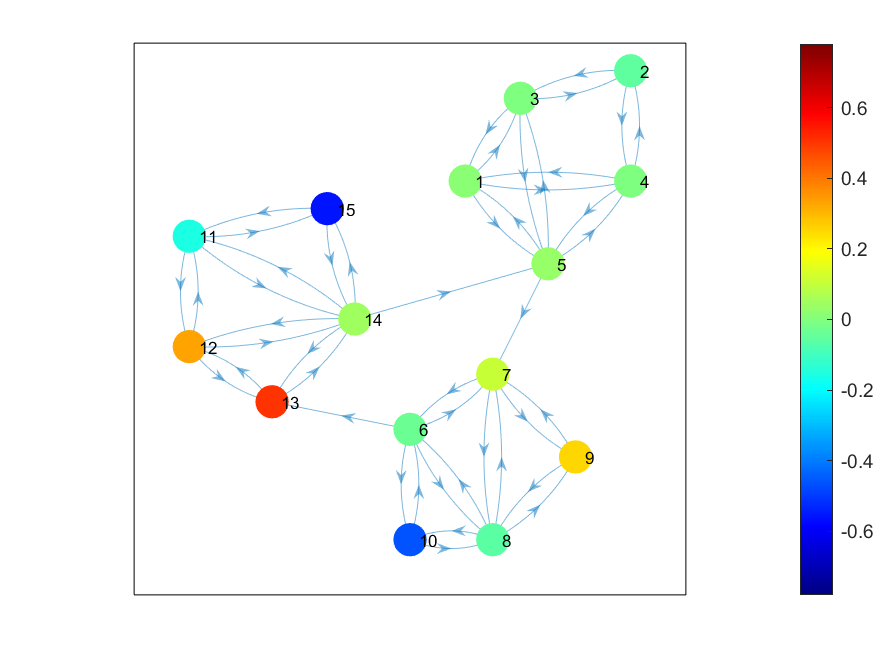} &
		\includegraphics[width=0.3\textwidth]{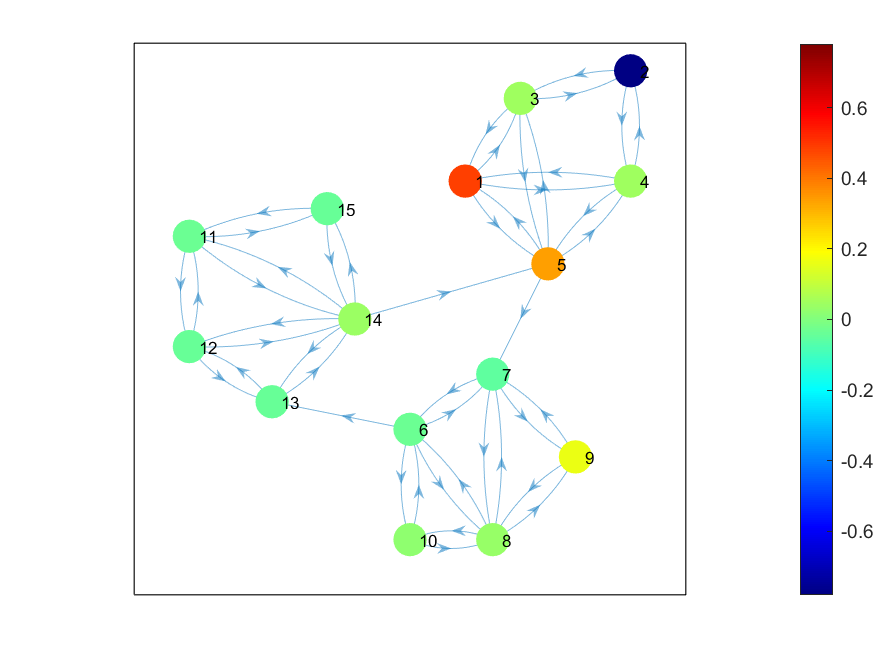} \\
		\includegraphics[width=0.3\textwidth]{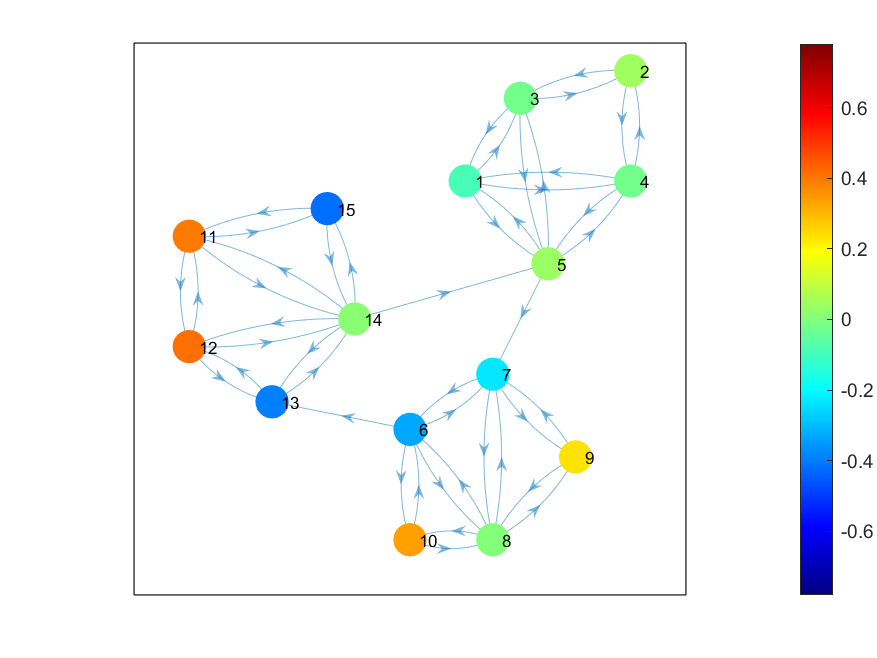} & 
		\includegraphics[width=0.3\textwidth]{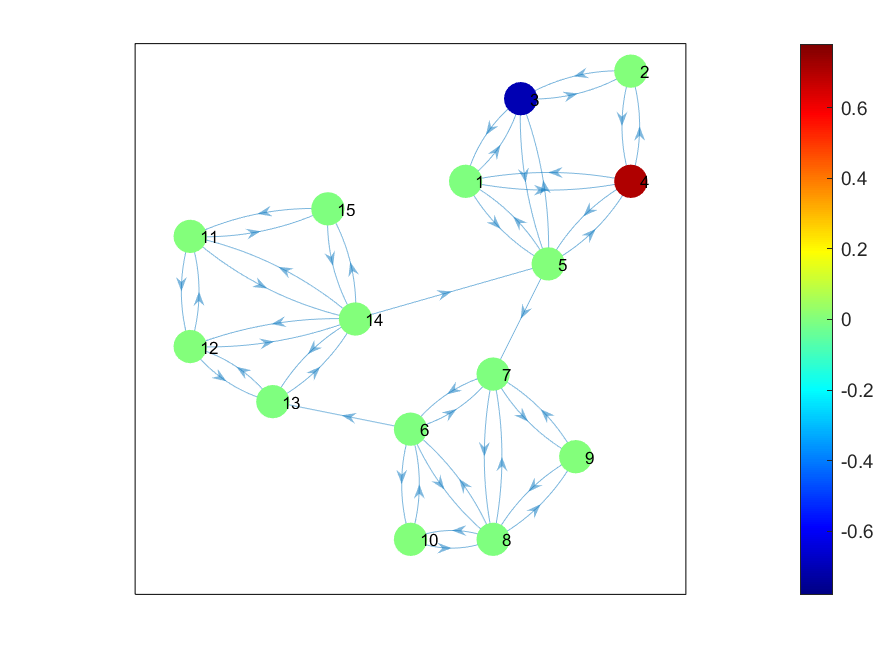} &
		\includegraphics[width=0.3\textwidth]{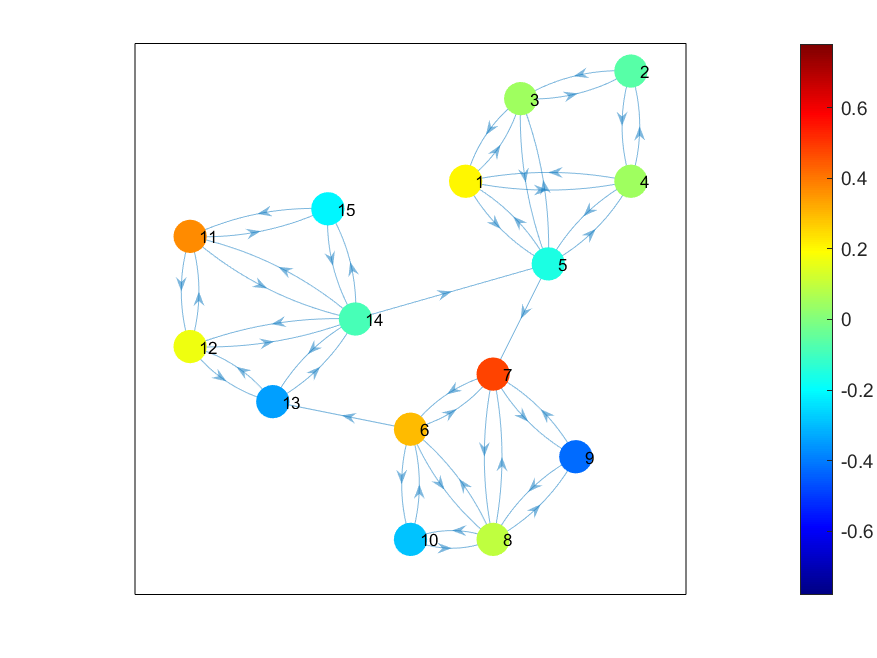} \\
		\includegraphics[width=0.3\textwidth]{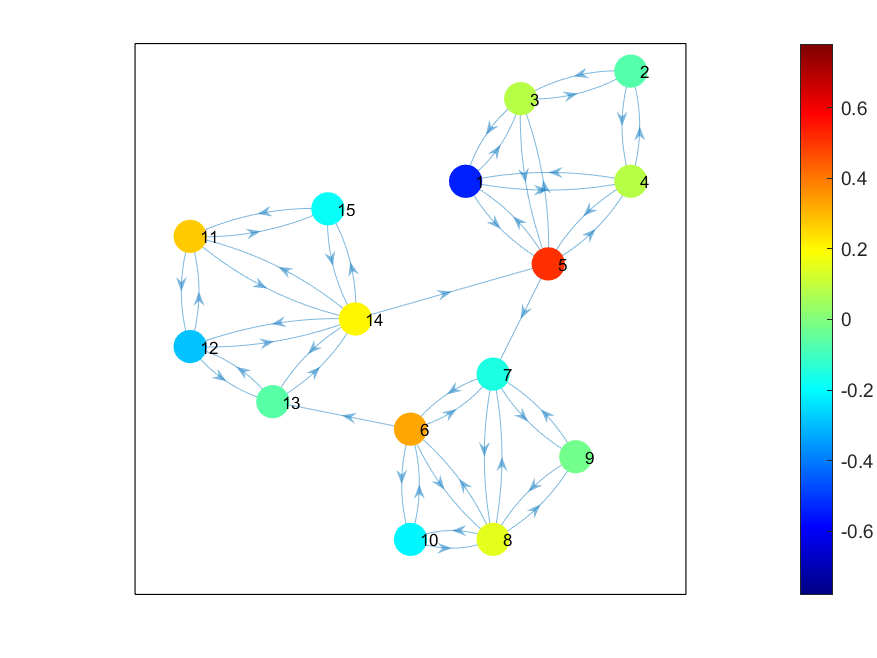} & 
		\includegraphics[width=0.3\textwidth]{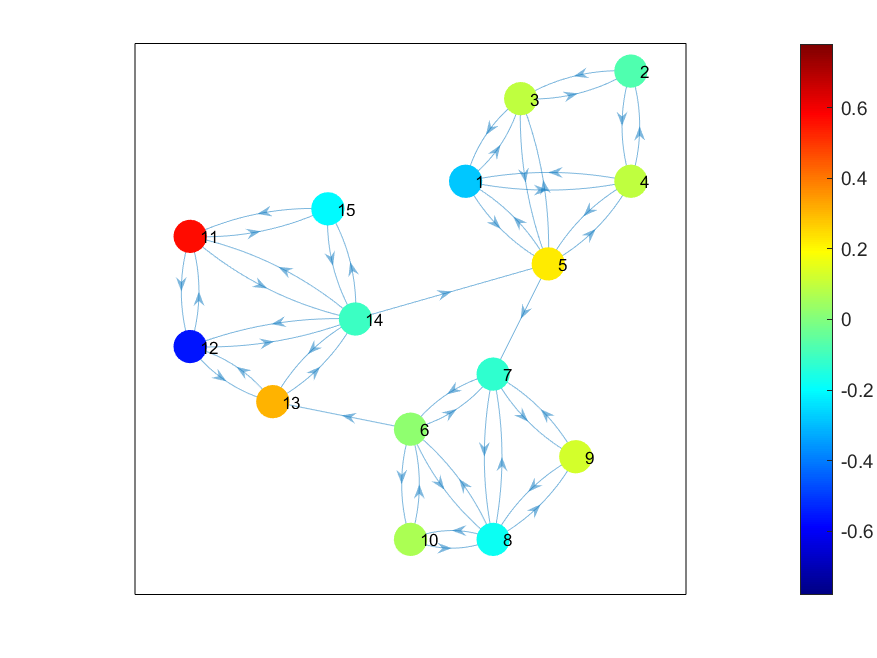} &
		\includegraphics[width=0.3\textwidth]{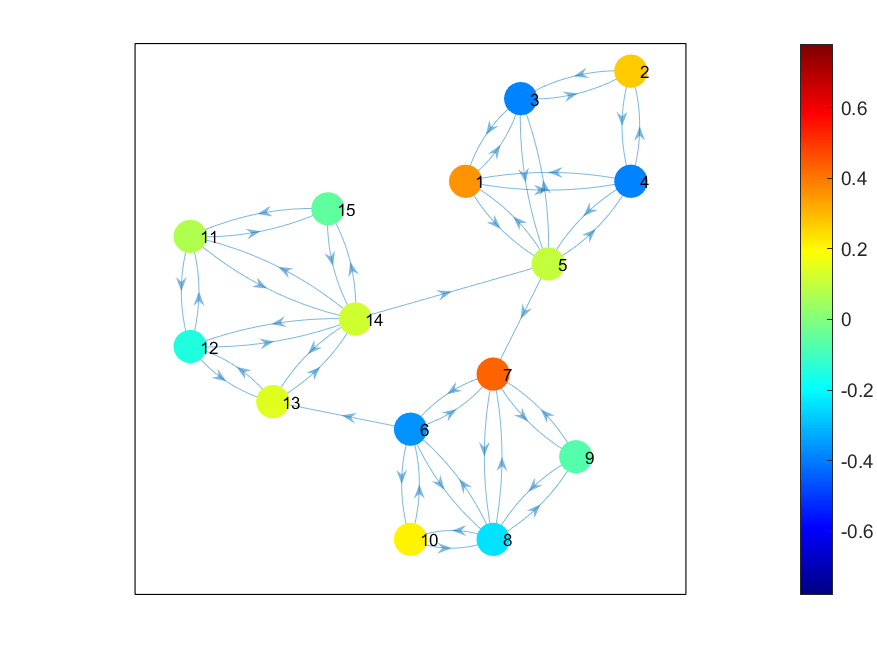} \\
		\includegraphics[width=0.3\textwidth]{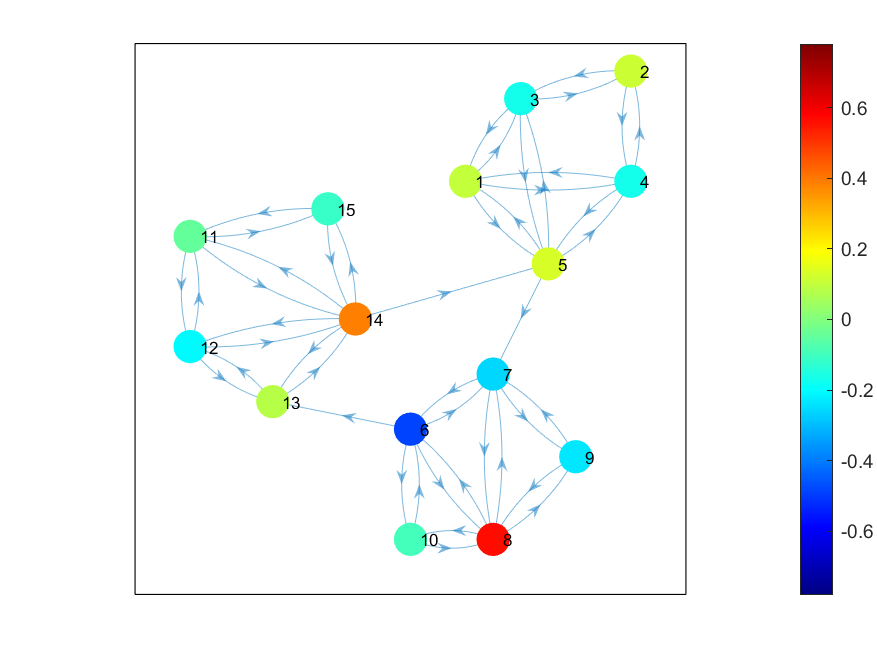} & 
		\includegraphics[width=0.3\textwidth]{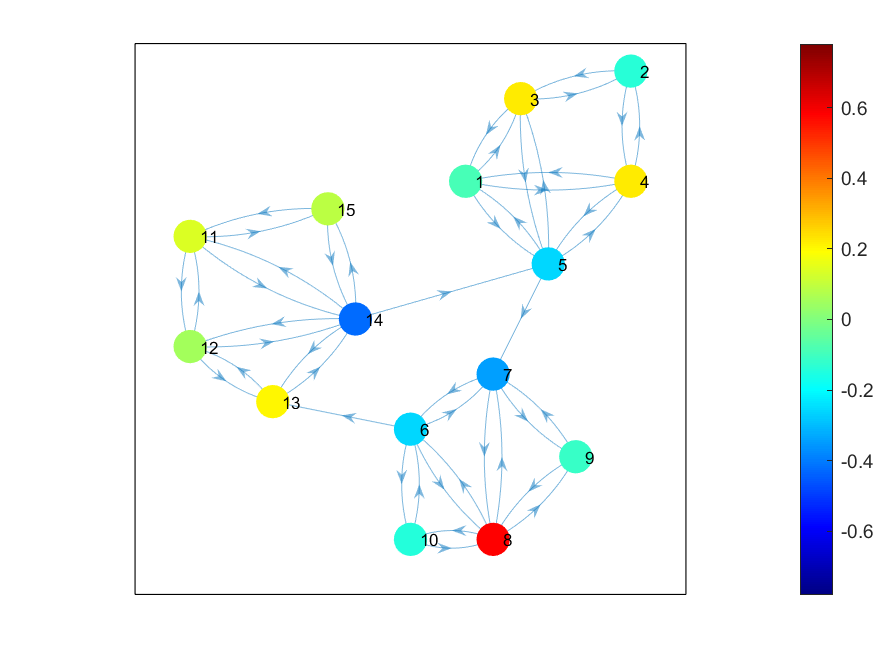} &
		\includegraphics[width=0.3\textwidth]{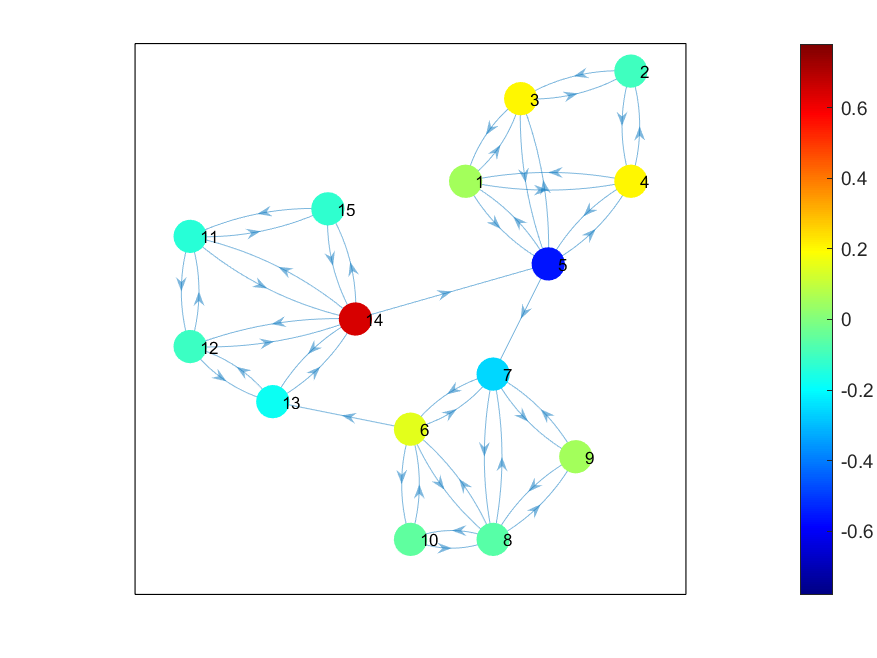}
		\et
		\caption{Eigenvectors of graph G15c.}
		\label{fig:eigv_15_c}
	\end{figure*}

Figures \ref{fig:eigv_15_a} and \ref{fig:eigv_15_c} show the values of the $N=15$ eigenvectors of the HaarD-Laplacian GFT of the G15a and G15c graphs discussed in Example \ref{ex:sarde}.
In subfigure $k$ (counting from the top, on rows), the coefficients of eigenvector
$u_k$ (sorted in increasing order of frequencies) are represented with colors;
since the coefficients are complex, we chose to represent the value
$\text{sign}(\text{Re} \, \tilde{u}_{\ell k}) \cdot |\tilde{u}_{\ell k}|$
associated with node $\ell$.

We note that the eigenvectors appear to follow the intuition on what can represent
low and high frequencies in such graphs.
The first eigenvector has nearly constant values.
The next few ones clearly respect the cluster structure, especially in the case of
G15a (Fig. \ref{fig:eigv_15_a}), which has the least connectivity between clusters.
The eigenvectors corresponding to high frequencies show significant larger variation in the
elements, as they should.
The nodes connected with other clusters are especially characterized by large coefficients,
more so in the graph containing a cycle, G15c (Fig. \ref{fig:eigv_15_c});
there, the last two eigenvectors show significant variations of the values corresponding
to nodes belonging to the cycle.
	
\bibliographystyle{elsarticle-num-names}
\bibliography{graph}

\end{document}